\PassOptionsToPackage{table}{xcolor}
\documentclass[11pt,letterpaper]{cmu} %
\usepackage[comma,numbers,sort,compress]{natbib}

\usepackage{amsmath,amsfonts,bm}

\def\eqref#1{equation~\ref{#1}}

\def\1{\bm{1}}

\DeclareMathAlphabet{\mathsfit}{\encodingdefault}{\sfdefault}{m}{sl}
\SetMathAlphabet{\mathsfit}{bold}{\encodingdefault}{\sfdefault}{bx}{n}

\usepackage{url}

\hypersetup{colorlinks = true,
            linkcolor = magenta,
            urlcolor  = brown,
            citecolor = brown,
            anchorcolor = magenta}

\usepackage{mathrsfs}
\usepackage{graphicx}
\usepackage{afterpage,float}
\usepackage{placeins}
\usepackage{booktabs}
\usepackage{tabularx}
\usepackage{subcaption}
\usepackage{amsmath}
\usepackage{amssymb}
\usepackage{mathtools}
\usepackage{amsthm}
\usepackage{wrapfig}
\usepackage{needspace}
\RequirePackage{algorithm}
\RequirePackage{algorithmic}
\definecolor{myblue}{RGB}{224, 240, 255} %
\usepackage[capitalize,noabbrev]{cleveref}

\theoremstyle{plain}

\theoremstyle{definition}

\theoremstyle{remark}

\usepackage[most,skins,theorems]{tcolorbox}
\tcbset{
  aibox/.style={
    width=\linewidth,
    top=7pt,
    bottom=2pt,
    colback=blue!6!white,
    colframe=black,
    colbacktitle=black,
    enhanced,
    center,
    attach boxed title to top left={yshift=-0.1in,xshift=0.15in},
    boxed title style={boxrule=0pt,colframe=white,},
  }
}
\newtcolorbox{AIbox}[2][]{aibox,title=#2,#1}
\definecolor{lightblue}{rgb}{0.22,0.45,0.70}%

\newtcolorbox{analysisbox}[1][]{
    enhanced jigsaw,
    colback=white,
    colframe=blue!75!black,
    fonttitle=\bfseries,
    boxsep=5pt,
    left=5pt,
    right=5pt,
    top=5pt,
    bottom=5pt,
    title=#1,
}
\newcommand{\methodname}{$\beta$-\textsc{Data}}
\title{Exploratory Midtraining for Robot Policies}

\author{
Kshitiz\textsuperscript{1, \#},
Zheyuan Hu\textsuperscript{1, \#},
Robyn Wu\textsuperscript{1},
Yizhou Li\textsuperscript{1},
Yunchu Zhang\textsuperscript{2},
Vattanary Tevy\textsuperscript{2},
Kavya Tayal\textsuperscript{2},
Shawn Yin\textsuperscript{2}, 
Aykut Onol\textsuperscript{3},
Kushal Arora\textsuperscript{3},
Dhruv Shah\textsuperscript{4}, 
Abhishek Gupta\textsuperscript{2},
Zackory Erickson\textsuperscript{1},
Aviral Kumar\textsuperscript{1} \vspace{2pt} \\
\textsuperscript{1}Carnegie Mellon University, 
\textsuperscript{2}University of Washington, 
\textsuperscript{3}Toyota Research Institute, 
\textsuperscript{4}Princeton University \vspace{4pt}\\
$^{\#}$\emph{Joint first authors}. Correspondence emails: \texttt{\{kshitiz2, zheyuanh\}@andrew.cmu.edu} \vspace{-4pt}
}

\begin{abstract}
\textbf{Abstract:} Robot policies predominantly learn from human-collected data emphasizing competent behavior, limiting coverage of the low-level behaviors needed to support autonomous improvement. We study how a form of midtraining can be used to build coverage over diverse low-level behaviors that connect challenging states of the robot and its environment to states from which further progress is possible (``connected behavioral coverage''). We also examine how this coverage supports downstream adaptation and online improvement. To this end, we introduce \methodname{}, a dataset collected via adversarial teleoperation, where a human adversary reversibly perturbs the environment while a human operator continues toward completion without resetting. We relabel trajectory segments with local outcomes in natural language and finetune a pretrained policy conditioned on recent action history. The resulting policy exhibits sustained use of more low-level behaviors. Across manipulation tasks in simulation and on three real-robot platforms, our approach yields more favorable data scaling laws, better zero-shot generalization, and more effective use of additional rollout time than training on standard demonstrations. During online finetuning, these policies need fewer rollouts to collect successful experience and reach higher success rates, suggesting that midtraining can prepare robots for both more robust task execution and continued autonomous improvement. Our project page and robot videos are available at \href{https://kshitiz-1225.github.io/exploratory-midtraining-robot-policies/}{https://kshitiz-1225.github.io/exploratory-midtraining-robot-policies/}.\looseness=-1
\end{abstract}

\begin{document}

\maketitle

\begin{figure}[H]
\centering
\vspace{-0.2cm}
\includegraphics[width=0.99\linewidth]{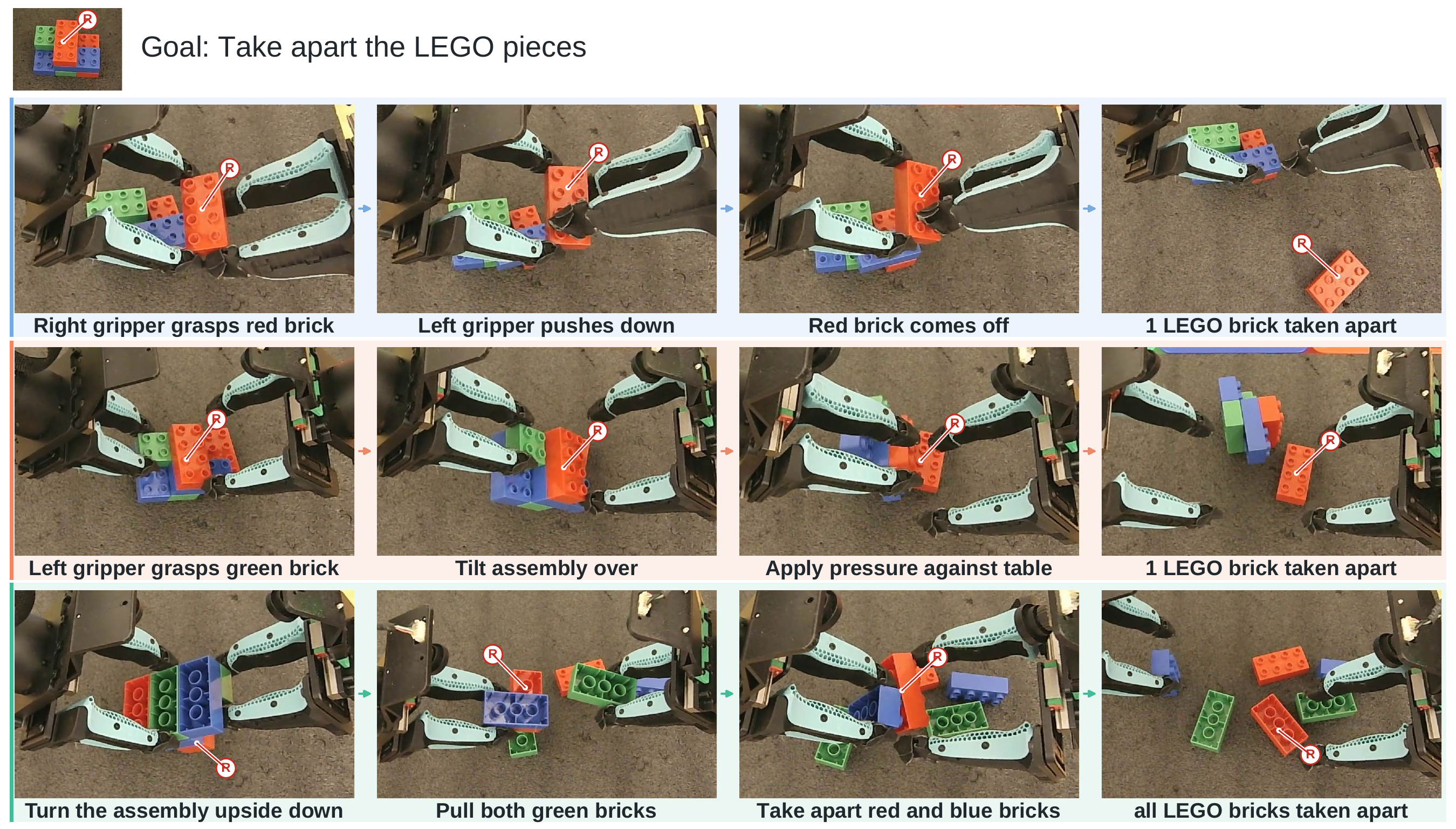}
\vspace{-0.25cm}
\caption{\footnotesize{\textbf{Connected behavioral coverage (CBC) induced by training on \methodname{} in LEGO disassembly.} From the same initial assembly and natural language goal, variations in grasps, orientations, and contacts produce diverse sequences of outcomes. The top and middle rollouts detach different red bricks, while the bottom rollout separates all bricks through a distinct strategy. The policy generates meaningful alternative interactions while sustaining progress toward task completion.}}
\vspace{-0.9cm}
\label{fig:lego-removal-routes}
\end{figure}
\clearpage

\vspace{-0.4cm}
\section{Introduction}
\label{sec:intro}
\vspace{-0.2cm}

Generalist robot policies are trained on demonstrations spanning diverse tasks and environments, but this diversity does not necessarily ensure coverage of the low-level behaviors needed to sustain progress in challenging configurations. Human-collected demonstrations, including teleoperation and UMI-style collection~\citep{chi2024universal}, largely consist of competent behaviors. While there are some natural mistakes and recovery segments in human data, the coverage over them depends on which difficulties arise during collection and how operators respond to them. As a result, collecting more demonstrations does not directly amplify the breadth or diversity of challenging conditions covered naturally by human data. We study whether deliberately broadening this coverage provides a more useful midtraining distribution. Our focus is on \emph{behavioral coverage}: increasing the coverage over low-level behaviors available to a policy for adapting, recovering, and continuing from the states encountered during execution. Such coverage can support both robust task completion and online improvement, since the behaviors a policy can execute also determine the experience it can generate for subsequent learning.

For instance, consider packing unfamiliar objects of different shapes, sizes, and rigidity into a box. Even with a correct high-level language plan specifying which object to manipulate next, the required low-level behavior depends on the evolving geometry of the task. Placing an object may require approaching from another direction, changing its grasp or orientation, repositioning it before insertion, or rearranging the existing contents of the box. Language can describe some of these strategies coarsely, but the policy must still possess the motions needed to execute them under the current physical constraints. The challenge is not primarily one of specifying \emph{what} should happen next, but of providing sufficient coverage over \emph{how} that outcome can be physically realized\footnote{To emphasize the need for low-level behaviors on our tasks, we evaluate GPT-6.2 Astra directly as the robot policy and find that it struggles to complete our LEGO disassembly task even when given far more time (Section~\ref{sec:zero-shot-tasks} and Appendix~\ref{app:frontier}). We often observe correct high-level planning from this reasoning model but not the correct low-level behavior.}. These realizations are difficult to enumerate precisely in advance, yet they determine whether the policy can sustain progress in unfamiliar configurations. Figure~\ref{fig:lego-removal-routes} illustrates this diversity of low-level behaviors in the disassembly of brick structures. 

Deliberately broadening coverage in robot data requires reaching challenging states and eliciting useful behaviors from them. Existing approaches often provide limited direct control over which challenging states are covered. Intervention-based methods~\citep{8793698,liu2022robot} collect human corrections during policy execution, but the resulting coverage is tied to the states visited and mistakes made by a particular policy, and may not carry over when the policy changes. Noise injection into the policy~\citep{fujimoto2018addressing} or teleoperation interface~\citep{laskey2017dart,zhang2025action} broadens experience by perturbing actions, but offers limited control over semantically useful challenging states. Language-based metadata conditioning~\citep{intelligence2026pi} can steer behavior along predefined axes, but does not by itself ensure coverage of the low-level behaviors needed from different states. These considerations motivate a collection procedure that deliberately drives the data collection system to challenging states and allows a competent operator to find coherent ways to adapt their demonstration and resume progress.

\emph{\textbf{How can we enhance low-level behavioral coverage?}} We seek data that deliberately broadens the useful behaviors available in challenging configurations of the system. Diversity over states or actions alone does not ensure this: a dataset may contain many states and many different individual actions at these states in disconnected ``islands''. What influences success in challenging situations is coverage over behaviors that can help move the robot from a challenging state to one from which further progress is possible. For example, when an object cannot be inserted into a crowded container, the robot may need to reorient it or push another object aside before inserting it (Figures~\ref{fig:sim-task11-recovery} and~\ref{fig:real-xarm-jar}). The relevant coverage includes these alternative behaviors together with the ``connective'' transitions needed to enable continued progress. We refer to this property as \emph{connected behavioral coverage}, and define it formally in Section~\ref{sec:cbc}.

To amplify connected behavioral coverage, we utilize \emph{adversarial data collection}~\citep{huang2025adversarial}, a protocol for human teleoperation, which separates data collection into two roles: an \emph{operator}, who attempts to complete the task, and an \emph{adversary}, who reversibly changes the environment or task to elicit adaptation. The adversary controls \emph{where} alternative behavior is needed, while the operator determines \emph{how} to respond. During packing, for example, the adversary may rotate an object, obstruct a placement, or change container occupancy, prompting the operator to reorient objects, rearrange the scene, or change their motion sequence. These interventions create additional challenging configurations, including outcomes of mistakes that the operator need not actually make. The operator continues from each perturbed state without resetting, providing trajectories that connect adaptation and recovery to resumed task progress. Since training on these trajectories is expected to produce more exploratory behaviors, we call the collected data \methodname{} (\textsc{Beta-Data}): \textbf{B}uilding \textbf{E}xploratory policies via mid\textbf{T}r\textbf{A}ining. 

We use \methodname{} to \emph{midtrain} a pretrained robot policy, expanding its behavioral repertoire before task-specific post-training. Long rollouts present two challenges: an episode-level instruction may not capture the local outcome being pursued, and the appropriate low-level continuation may depend on previous actions. We address these challenges by relabeling coherent trajectory segments with their local outcomes and conditioning the policy on recent action history. The annotations specify what to accomplish locally, while the history provides context for selecting how to continue.

\textbf{Results.} We evaluate the utility of midtraining on \methodname{} through comprehensive data-scaling studies and downstream online improvement. On a long-horizon bimanual simulated packing environment with twenty evaluation tasks, eight in-distribution (ID) and twelve out-of-distribution (OOD), \methodname{} improves the success rate of a $\pi_{0.5}$~\citep{intelligence2025pi_} base pretrained policy from 54.7\% with standard demonstrations to 69.7\% with \methodname{}. The gains remain large on OOD tasks with long time horizons, where success improves from 39.2\% to 57.6\%. Analysis of behavior associates these gains with productive retries and sustained progress through the final stages of a task. We further find more favorable data-scaling curves with \methodname{}, and observe similar improvements across three real-world long-horizon bimanual tasks involving item packing, disassembling complex shapes, and folding clothes on three different robot platforms. We also find that increasing the time limit of the rollout increases performance of the policy trained with \methodname{} further. We also show that policies trained with \methodname{} are better able to collect more useful and larger amounts of successful experience for online improvement via practice. Finally, we compare against a frontier model GPT-6.2 Astra and find that it plans correctly but struggles with low-level execution.

\vspace{-0.2cm}
\section{Preliminaries and Notation}
\label{sec:prelim}
\vspace{-0.2cm}

We consider solving tasks specified in natural language $l\in\mathcal L$, with the robot low-level action space denoted by $\mathcal A$. At timestep $t$, the robot receives an observation $s_t$, consisting of image observations and proprioceptive states, and executes an action $a_t\in\mathcal A$.
The policy predicts an action chunk
$a_{t:t+H-1} = (a_t,\ldots,a_{t+H-1})$ of $H$ actions rather than a single action.
Because many behaviors of interest in this work depend on previous attempts and failures, we train history-conditioned policies.
We denote the action history at timestep $t$ by $h_t = (a_{0:t-1})$,
and write a policy with learned parameters $\theta$ as $a_{t:t+H-1}\sim\pi_\theta(\cdot|s_t,h_t,l)$.
Our implementation uses the $W=120$ most recent actions, $\bar h_t=(a_{\max(0,t-W)},\ldots,a_{t-1})$, expressed relative to the current state (Section~\ref{sec:method}). 

Executing a policy for $T$ timesteps produces a trajectory $\tau=(s_0,a_0,\ldots,s_T)$. Each task $l$ induces a binary success function
$r_l(\tau) \in \{0,1\}$ indicating completion at the end of the rollout. We also measure task progress, a ``dense'' metric quantifying the amount of progress made by the rollout trajectory (Appendix~\ref{app:task-progress}). Imitation learning trains $\pi_\theta$ on a dataset
$\mathcal{D}_{\mathrm{demo}}$
of demonstrations.
A demonstration can be written as $(\tau_i,l_i)$ with $\tau_i = (s^i_0,a^i_0,\ldots,s^i_{T_i})$, 
where the human controls the robot toward successful completion of instruction $l_i$.
Such trajectories may be collected through teleoperation or other forms of human-generated behavior data, but in each case predominantly consist of goal-directed behavior produced by competent human operators.
For a dataset $\mathcal D$ of such demonstrations, we write the log-loss form of behavioral cloning as $\mathcal{L}_{\mathrm{BC}}(\theta;\mathcal{D})
    =
    -\mathbb{E}_{(\tau,l)\sim\mathcal{D}}
    \left[
        \sum_t
        \log \pi_\theta
        (a_{t:t+H-1} | s_t, h_t, \tilde l_t)
    \right].$
For \textbf{subtask-conditioned} training, $\tilde l_t$ is the local-outcome annotation of the segment containing timestep $t$; for full-task conditioning, $\tilde l_t=l$ throughout the rollout. The log-loss expression describes behavioral cloning conceptually; our implementation on top of a flow-matching pretrained policy instead minimizes its standard conditional flow-matching objective. 
In this work, we ask how data should be collected so that a robot policy acquires broad coverage over low-level behaviors, rather than primarily reproducing expert execution.

\textbf{Online finetuning.}
Policies that exhibit broader behavioral coverage should provide a better initialization for autonomous online improvement on previously unseen tasks.
Given a downstream task $l^\star$ and an initial policy $\pi_{\theta_0}$, we perform filtered behavioral cloning (filtered BC) over successive rounds.
At round $k$, we roll out the current policy $\pi_{\theta_k}$ on $l^\star$ until it produces $n_s$ successful trajectories $\tau_k^{(1)}, \cdots, \tau_k^{(n_s)}$.
We add these trajectories to an online success buffer, $\mathcal{D}^{(k)}_{\mathrm{on}}
    =
    \mathcal{D}^{(k-1)}_{\mathrm{on}}
    \cup \{\tau_k^{(1)}, \cdots, \tau_k^{(n_s)}\},
$
and update the policy using behavioral cloning. Following standard practice, we retain a fraction of the original offline data during each update~\citep{kowshik2026adaptation,mark2024policy} to regularize the policy and mitigate forgetting. We describe our procedure for online finetuning in Section~\ref{sec:online-finetuning}, with simulation replay and optimization details in Appendices~\ref{app:training-settings} and~\ref{app:online-task-curves}, and the real-world protocol in Appendix~\ref{app:real-online}.

\vspace{-0.2cm}
\section{Environments, Tasks, and Robot Setups}
\label{sec:envs_tasks}
\vspace{-0.2cm}

\begin{wrapfigure}{r}{0.4\linewidth}
\vspace{-0.7cm}
\centering
\includegraphics[width=\linewidth]{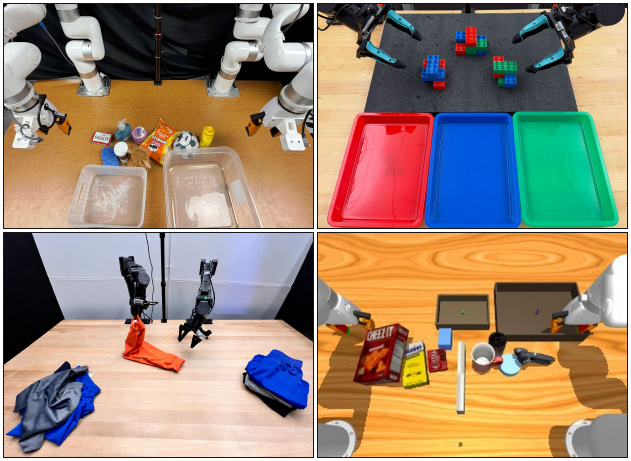}
\vspace{-0.6cm}
\caption{\footnotesize{\textbf{Robot setups \& tasks.} From L to R, top to bottom: real-world grocery packing on bimanual xArm 7, LEGO disassembly and sorting on bimanual YAM, garment folding on Raiden bimanual YAM, and simulated grocery packing in MuJoCo.}}
\label{fig:task-overview}
\vspace{-0.4cm}
\end{wrapfigure}
We study several long-horizon bimanual manipulation tasks that require complex low-level skills across a sequence of dependent subtasks. To study coverage over low-level behavior, we choose tasks that remain challenging even when a capable high-level planner specifies \emph{what} to do next, including a rearrangement or a retry. The robot must still determine \emph{how} to grasp, reorient, and move objects to realize that outcome under the constraints of the current scene. For example, placing an object in a crowded container may require changing the grasp or moving an obstruction before attempting the placement. These decisions also affect subsequent subtasks: an early placement can restrict later motions, and execution errors can accumulate over long horizons. Completing the task, therefore, requires the policy to adapt its low-level behavior reactively and sustain progress via exploration.

\textbf{Simulation and real-world tasks.} We instantiate these settings in one simulated and three real-world domains. For simulation, in MuJoCo~\citep{todorov2012mujoco}, we construct 72 distinct \emph{grocery packing} training tasks and 12 held-out tasks, in which bimanual xArm 7 robot arms arrange multiple grocery items of varying shapes and sizes between differently sized containers, posing non-trivial spatial constraints on the low-level dexterous skills as shown in Figure~\ref{fig:task-overview}. We describe the twenty evaluation tasks, their split into in-distribution and held-out tasks, task goals, numbers of subtasks, and horizon length in Table~\ref{tab:packing-task-inventory}~of~Appendix~\ref{app:packing-tasks}.

Our real-world experiments are conducted on three bimanual robot platforms: \textbf{(1)} \textbf{\emph{grocery packing}} on a bimanual xArm 7 robot, where grocery items are to be packed into two containers with spatial constraints; \textbf{(2)} \textbf{\emph{LEGO disassembly}} on a bimanual YAM robot, where complex LEGO brick structures are taken apart and sorted into colored trays, requiring precise contact forces and dexterous manipulation strategies; and \textbf{(3)} \textbf{\emph{garment folding}} on Raiden~\citep{raiden2026} bimanual YAM arms, where pants and T-shirts are folded.  Detailed task definitions are in Appendix~\ref{app:task-progress}, and Appendix~\ref{app:robot-setup} describes each robot platform.

\textbf{Full tasks and subtasks.}
A \emph{full task} specifies the desired final state, including object destinations and constraints or objectives (e.g., ``pack all objects efficiently'', ``fold the shirt''). It may require a sequence of \emph{subtasks} $\tilde l$, each specifying a local semantic outcome, such as ``pack the sugar box into the medium tray'' or ``lift up the bottom left of the shirt''. Neither level fully specifies the \emph{low-level behavior}: the sequence of motor actions used to achieve that outcome. Depending on the scene, the same subtask may require different grasps, approach directions, reorientations, bimanual transfers, temporary restaging, or repeated attempts. Our goal is to broaden coverage over these low-level realizations, as opposed to only increasing the diversity of subtasks or improving language subtask following.

\textbf{Hierarchical and direct execution.}
In the \emph{hierarchical} setting, a high-level planner provides the current subtask instruction $\tilde l_t$, and the policy produces continuous actions to realize it. In the \emph{direct} setting, the policy receives the full task instruction $l$ throughout the rollout, without a high-level planner. The hierarchical setting tests whether broader behavioral coverage helps even when \emph{what} to do next is given, leaving the policy to determine \emph{how} to achieve it.

\vspace{-0.3cm}
\section{\methodname{}: Exploratory Policies via Adversarial Midtraining}
\label{sec:method}
\vspace{-0.2cm}

We introduce \methodname{}, a midtraining dataset collected to broaden the low-level behavioral coverage of pretrained robot policies. We first characterize the notion of this coverage, then describe our data collection protocol, and finally our training procedure.

\vspace{-0.2cm}
\subsection{Connected Behavioral Coverage}
\label{sec:cbc}
\vspace{-0.1cm}

Na\"ively diversifying initial states during teleoperation can expose operators to challenging configurations and elicit recovery behaviors. However, this alone does not ensure broad coverage of the configurations robot policies might encounter during execution, and the various low-level action sequences needed to recover and resume progress from them. Directly diversifying teleoperator actions by injecting noise into the interface or changing the instructions given to the operator might help reach some of these challenging configurations, but still offers limited control over the set of states where multiple low-level behaviors should be illustrated. In addition, adding action noise may reduce the overall coherence of the learned policy. To circumvent these limitations, we seek high \emph{connected behavioral coverage}: coverage over challenging states we expect to see when a policy is deployed, together with diverse low-level continuations that connect them to states from which further progress is possible.

\setlength{\emergencystretch}{1em}\textbf{Quantifying connected behavioral coverage (CBC).} We first define connected behavioral coverage 
quantitatively to enable comparisons between datasets. Conceptually, a dataset of episodes can be represented as a graph, where nodes correspond to abstract states of the environment and robot, and directed edges represent transitions demonstrated in the dataset. The number of distinct edges then provides a measure of CBC. To instantiate this idea, we annotate existing robot data. Each annotated trajectory segment manipulates one object, taking it from a starting abstract state to an ending abstract 
\begin{wrapfigure}{r}{0.6\linewidth}
\vspace{-0.15cm}
\centering
\includegraphics[width=0.99\linewidth]{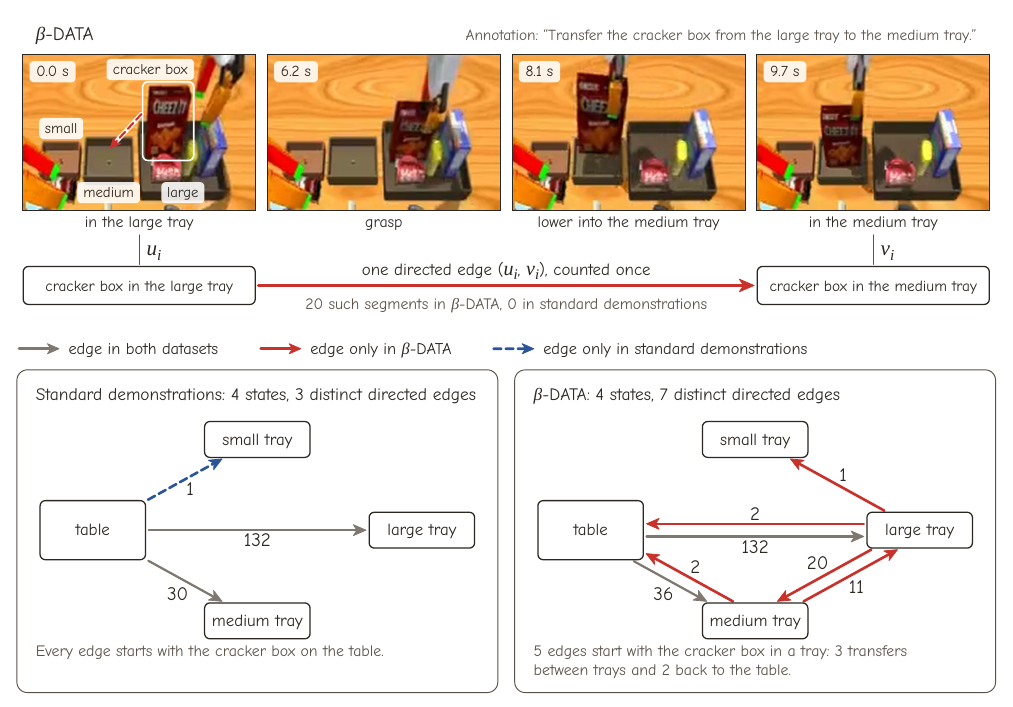}
\vspace{-0.2cm}
\caption{\footnotesize{\textbf{Measuring connected behavioral coverage for one object.} \textit{\textbf{Top:}} one annotated trajectory segment in \methodname{} transfers the cracker box from the large tray to the medium tray and contributes one directed edge between the two object-location states. \textit{\textbf{Bottom:}} all directed edges of the cracker box in simulated packing. The number on an edge is the number of annotated trajectory segments that demonstrate it. Standard demonstrations and \methodname{} contain the same four states. Standard demonstrations connect them with three distinct directed edges, and \methodname{} connects them with~seven.}}
\label{fig:cbc-example}
\vspace{-0.4cm}
\end{wrapfigure}
state. When the two states differ, the segment demonstrates a transition between them, which we count as one directed edge from the first to the second. We measure the connected behavioral coverage of a dataset by the number of distinct directed edges in its demonstrations at a given data budget. Figure~\ref{fig:cbc-example} illustrates this count for the cracker box in simulated packing. A transfer from the large tray to the medium tray contributes one edge, regardless of how often it occurs. Both datasets contain the box on the table and in each of the three trays, but standard demonstrations connect these four states with three distinct directed edges, while \methodname{} connects them with seven.

\begin{wrapfigure}{r}{0.5\linewidth}
\vspace{-0.1cm}
\centering
\includegraphics[width=\linewidth]{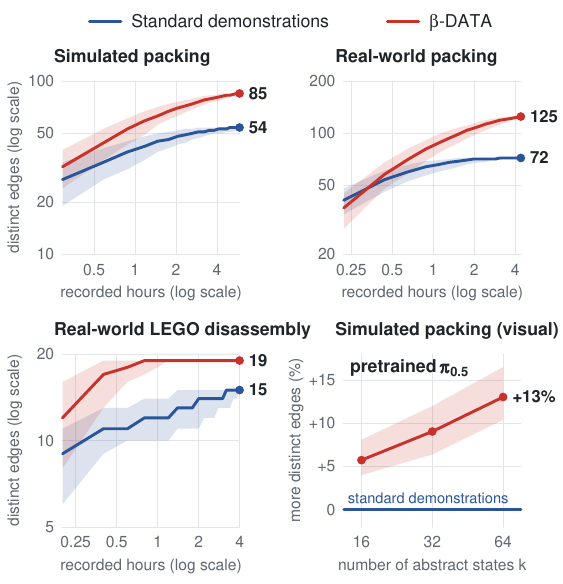}
\vspace{-0.65cm}
\caption{\footnotesize{\textbf{\methodname{} contains more distinct directed edges at matched data budgets.} Number of distinct directed edges (Eq.~\ref{eq:cbc}) as a function of the data budget in recorded hours. We observe that CBC is higher with \methodname{} and grows more rapidly with the amount of data in both packing domains, as indicated by steeper log-log slopes (0.30 vs. 0.22 in simulation and 0.37 vs. 0.15 in the real world). \textit{\textbf{Bottom right:}} percentage increase in distinct directed transitions over standard demonstrations at a matched data budget in simulated packing, as a function of the number of clusters $k$ of pretrained $\pi_{0.5}$ vision features. Shading shows the 2.5 to 97.5 percentile range over 1,000 random selections of demonstrations in the first three panels, and a 95\% bootstrap interval in the bottom-right panel.}}
\vspace{-0.9cm}
\label{fig:cbc-budget}\label{fig:cbc-latent}
\end{wrapfigure}

\setlength{\emergencystretch}{0pt}Formally, suppose we have access to a state representation $\phi$ that maps observations to a discrete set of abstract states. We use the same representation and segmentation rule when comparing datasets. For each annotated trajectory segment $\tau_i$, let $u_i=\phi(s_{b_i}), v_i=\phi(s_{e_i})$,
where $s_{b_i}$ and $s_{e_i}$ are the starting and ending observations of the segment. Let $\mathcal D_B$ denote the demonstrations selected under a data budget $B$, measured in recorded hours, and let $\mathcal I(\mathcal D_B)$ index their annotated trajectory segments. We define
\begin{equation}
\begin{aligned}
E_{\mathcal D}(B)
&=\{(u_i,v_i): i\in\mathcal I(\mathcal D_B),\ u_i\neq v_i\},\\
\mathrm{CBC}(\mathcal D;B)
&=|E_{\mathcal D}(B)|.
\end{aligned}
\label{eq:cbc}
\end{equation}
This definition counts the distinct directed transitions demonstrated between abstract states. Each distinct directed edge counts once, regardless of how often it appears or which low-level action sequences produce it. We use an object-based discrete representation $\phi(s,o_i)$ that maps the tracked object $o_i$ and its location in observation $s$ to an abstract state in Figure~\ref{fig:cbc-budget} (first three panels) and a discretized, feature-vector based representation for an observation $s$ as an abstract state in Figure~\ref{fig:cbc-budget} (bottom right). We describe our procedure for counting edges in Appendix~\ref{app:cbc-measurement}.

\textbf{Measuring connected behavioral coverage.} Using Equation~\ref{eq:cbc}, we compare the distinct directed edges in the annotated demonstrations at matched data budgets (Figure~\ref{fig:cbc-budget}). \methodname{} contains 85 versus 54 edges for standard demonstrations in simulated packing, 125 versus 72 for standard demonstrations in real-world packing, and 19 versus 15 in LEGO disassembly. Even when we keep the abstract states the 
same, by restricting both real-world packing graphs to the 82 object-location states that both datasets contain, \methodname{} still has 121 distinct edges versus 72 for standard demonstrations, a 68\% increase. Thus, \methodname{} adds new transitions between the same abstract states, and not only new states.

\textbf{Measuring CBC with pretrained visual features.} In simulated packing, we embed the start and end observations of each annotated segment with the frozen vision encoder of pretrained $\pi_{0.5}$, cluster the 
embeddings from both datasets together, and count distinct directed transitions between clusters using Equation~\ref{eq:cbc}. Both datasets visit every cluster, yet \methodname{} connects more of these states at the matched data budget, and the difference grows with the number of clusters (Figure~\ref{fig:cbc-latent}, bottom right). Appendix~\ref{app:cbc-measurement} gives the details.

\vspace{-0.2cm}
\subsection{Collecting High CBC Data \& Midtraining}
\vspace{-0.1cm}

\begin{figure}[t]
\centering
\vspace{-0.2cm}
\includegraphics[width=\linewidth]{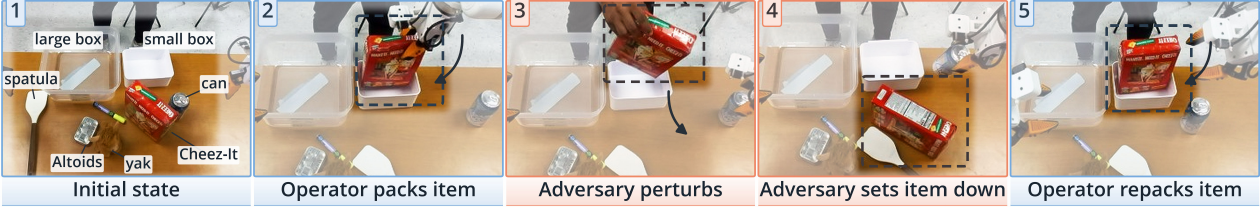}
\vspace{-0.7cm}
\caption{\footnotesize{\textbf{Perturbation and recovery during the adversarial data collection of \methodname{}.} The operator packs an item, the adversary returns it to the table, and the operator repacks it. We use this protocol to collect \methodname{}. Figure~\ref{fig:real-collection-recovery} in Appendix~\ref{app:real-xarm-rollouts} shows more of this episode.}}
\label{fig:collection-recovery-example}
\vspace{-0.4cm}
\end{figure}

\textbf{Protocol: \emph{operator-adversary data collection (Figure~\ref{fig:collection-recovery-example}).}}
To attain connected behavioral coverage, we utilize the adversarial data collection protocol of \citet{huang2025adversarial} since our goal is to primarily understand if CBC is important for autonomous improvement. In simulation, the adversary is implemented programmatically. During real-world collection, two humans jointly collect each episode: an \emph{operator} continually attempts to complete the task, while an \emph{adversary} intervenes during the episode to change the environment or task instructions and force the operator to adapt. This separation allows the adversary to control \emph{where} alternative behavior is elicited while leaving \emph{how} to respond up to the operator. The objective of the operator remains task completion, and they are never instructed to behave diversely or produce a particular recovery strategy, avoiding arbitrary variation in low-level actions at states where expert behavior is already sufficient. After each intervention, the operator continues from the perturbed state without restarting. As a result, the collected episodes are long-horizon and contain connected sequences from disruption through adaptation to resumed task progress, providing alternative low-level continuations precisely at critical states. We refer to the collected data as \methodname{}. Appendices~\ref{app:sim-packing-filmstrips}, \ref{app:real-xarm-rollouts}, and~\ref{app:lego-filmstrips} show examples of this protocol in simulated and real-world packing, and LEGO disassembly.

\textbf{Midtraining on the collected data.}
The protocol above produces distinct behaviors as the operator adapts to changing states and goals. Conditioning every action in such a trajectory only on the original task instruction can present a misspecified setting: the operator may locally remove an obstruction, transfer an object, retry a failed maneuver, or temporarily undo progress before continuing toward the final goal. To train on \methodname{}, we incorporate two design choices. First, \textcolor{magenta}{\textbf{D1: segment-level outcome relabeling}}, where we segment each episode according to the local outcome achieved and annotate each segment with this outcome in language, $\tilde l_t$. The annotation specifies \emph{what} the operator is trying to accomplish locally, while the demonstrated actions determine \emph{how} that outcome is realized. We apply the same relabeling procedure to trajectories from standard demonstrations. Second, \textcolor{magenta}{\textbf{D2: action-history conditioning}}, where we condition the policy on a short history of recent actions. At critical states, the same observation and subtask instruction can require different continuations depending on whether the robot has just approached an object, failed a grasp, withdrawn, or begun a retry. Recent actions therefore provide \emph{behavioral context} for selecting among these continuations, rather than merely compensating for partial observability. We then train the pretrained robot policy to predict future action chunks conditioned on the current observation, segment-level language annotation, and recent action history. At evaluation, a high-level planner supplies a corresponding subtask instruction in the hierarchical setting described in Section~\ref{sec:envs_tasks}. We ablate these design choices in Section~\ref{sec:zero-shot-tasks}.

\textbf{Implementation details.}
We midtrain the open-source vision-language-action model $\pi_{0.5}$~\citep{intelligence2025pi_}. To implement action-history conditioning, we express each of the previous 120 actions (two seconds at 60\,Hz, or four seconds for garment folding at 30\,Hz) relative to the current state, keeping the gripper command absolute, and encode them into four tokens that are provided as a prefix to the action expert. During training, we add small Gaussian noise to the history and drop the whole history with probability 0.2. During full-data midtraining, we fully finetune the image encoder and action expert and train rank-16 LoRA adapters~\citep{hu2021lora} on the language backbone, using the standard flow-matching objective for 10 epochs. Training configurations are listed in Appendix~\ref{app:training-settings}, and simulation evaluation details are provided in Appendix~\ref{app:zero-shot-simulation}.

\vspace{-0.2cm}
\section{Results: Zero-Shot Performance Scaling with \methodname{}}
\label{sec:zero-shot-tasks}
\vspace{-0.2cm}
\begin{figure}[t]
\centering
\vspace{-0.2cm}
\includegraphics[width=0.99\linewidth]{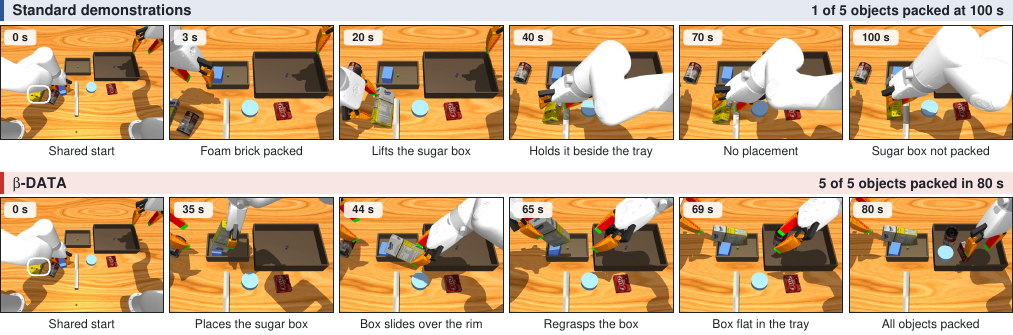}
\vspace{-0.25cm}
\caption{\footnotesize{\textbf{Recovery after a failed placement in simulated packing, from the same initial state.} On an OOD task, both policies start from the same state and are instructed to pack the sugar box into the medium tray. The policy trained on standard demonstrations (top) packs the foam brick, then lifts the sugar box and holds it beside the tray until the rollout ends, packing one of five objects. The policy trained on \methodname{} (bottom) places the sugar box, and when the box slides over the tray rim, it regrasps the box, sets it flat in the tray, and packs all five objects.}}
\label{fig:sim-task11-recovery}
\vspace{-0.3cm}
\end{figure}
We first study the zero-shot benefits of midtraining on \methodname{} by answering two questions. \textbf{RQ1:} Does \methodname{} improve connected behavioral coverage and improve zero-shot generalization? We evaluate task progress and success in simulation and the real world, examine low-level policy behavior, and test harder configurations to answer this question. \textbf{RQ2:} How do these benefits scale with training data (i.e., training compute) and rollout time (i.e., test-time compute)? We also ablate the two main design choices from Section~\ref{sec:method}: \textbf{Ablation 1:} action-history conditioning (\textcolor{magenta}{\textbf{D2}}), which provides behavioral context; and \textbf{Ablation 2:} segment-level outcome annotations (\textcolor{magenta}{\textbf{D1}}), which condition control on the local subtask.

\begin{wraptable}{r}{0.65\textwidth}
\vspace{-0.1cm}
\centering
\vspace{0.1cm}
\footnotesize
\setlength{\tabcolsep}{1.5pt}
\renewcommand{\arraystretch}{1.10}

\begin{tabularx}{\linewidth}{
@{}>{\raggedright\arraybackslash}p{0.27\linewidth}
@{\hspace{1pt}}
*{4}{>{\centering\arraybackslash}X}@{}
}
\toprule
& \multicolumn{2}{c}{\scriptsize\textbf{Success rate}}
& \multicolumn{2}{c}{\scriptsize\textbf{Progress (\%)}} \\
\cmidrule(lr){2-3}\cmidrule(lr){4-5}
{\scriptsize\textbf{Task}}
& {\scriptsize\textbf{Standard}}
& {\scriptsize\textbf{\methodname{} \textit{(Ours)}}}
& {\scriptsize\textbf{Standard}}
& {\scriptsize\textbf{\methodname{} \textit{(Ours)}}} \\
\midrule

\multicolumn{5}{@{}l}{\textit{Simulation}} \\

\hspace{0.6em}Short, OOD
& 136/150 & \textbf{148/150}
& 95.4 & \textbf{99.4} \\

\hspace{0.6em}Medium, ID
& 121/150 & \textbf{137/150}
& 88.4 & \textbf{94.3} \\

\hspace{0.6em}Medium, OOD
& 142/200 & \textbf{183/200}
& 84.9 & \textbf{94.7} \\

\hspace{0.6em}Long, ID
& 50/250 & \textbf{85/250}
& 41.4 & \textbf{58.9} \\

\hspace{0.6em}Long, OOD
& 98/250 & \textbf{144/250}
& 64.0 & \textbf{77.6} \\

\midrule
\multicolumn{5}{@{}l}{\textit{Real robot (OOD)}} \\

\hspace{0.6em}YAM, interlock
& 5/20 & \textbf{19/20}
& 43.3 & \textbf{96.7} \\

\hspace{0.6em}YAM, double
& 2/20 & \textbf{~6/20}
& 35.8 & \textbf{59.4} \\

\hspace{0.6em}xArm, packing
& 4/20 & \textbf{12/20}
& 45.0 & \textbf{74.4} \\

\hspace{0.6em}Raiden YAM, pants
& 3/20 & \textbf{14/20}
& 47.5 & \textbf{83.9} \\

\hspace{0.6em}Raiden YAM, T-shirt
& 5/20 & \textbf{~9/20}
& 63.9 & \textbf{76.7} \\

\bottomrule
\end{tabularx}
\vspace{-0.2cm}
\caption{\footnotesize
\textbf{\textit{Zero-shot evaluation on simulation and real-robot tasks.}}
Midtraining robot policies on \methodname{} yields higher success than standard demonstrations across all reported task groups, including challenging held-out configurations on real-robot platforms.
The largest gains in task progress occur on long-horizon tasks. For packing tasks, progress is the fraction of objects correctly placed out of the total number of objects. Real-robot progress follows the task-specific definitions in Appendix~\ref{app:task-progress}. Full simulation results are in Table~\ref{tab:zero-shot-subtask-detailed}.}
\vspace{-0.35cm}
\label{tab:zero-shot-subtask}
\end{wraptable}
\newcounter{resultnum}\refstepcounter{resultnum}\label{res:zero-shot}\textcolor{lightblue}{\textbf{Result~\theresultnum: Zero-shot successes from \methodname{} compared with standard demonstrations.}} We first study whether training on \methodname{} improves zero-shot success rates on held-out tasks. Table~\ref{tab:zero-shot-subtask} summarizes results for policies trained on standard demonstrations and \methodname{} on simulated and real-world tasks. The twenty simulated packing tasks consist of in-distribution (ID) tasks represented during training and out-of-distribution (OOD) tasks with held-out goals and configurations and are grouped into ``short'', ``medium'', and ``long'' tasks according to the task structure defined in Appendix~\ref{app:packing-tasks}. We provide a six-minute 
limit for any evaluation rollout in simulation and a five-minute limit for real-world xArm packing. We report detailed per-task results in Appendix~\ref{app:zero-shot-simulation}. Appendices~\ref{app:sim-packing-filmstrips}, \ref{app:real-xarm-rollouts}, and~\ref{app:lego-filmstrips} show selected rollouts of both policies in simulated packing, real-world packing, and LEGO disassembly. Overall, we see that \methodname{} substantially improves zero-shot task completion compared with training on standard demonstrations, both on the success rate of the entire task and task progress. 

\setlength{\emergencystretch}{0pt}\textbf{Comparison with PostBC~\citep{wagenmaker2025posteriorbehavioralcloningpretraining}.} We also compare with posterior behavioral cloning (PostBC), which attempts to broaden pretrained robot-policy behavior by modeling uncertainty in demonstrator actions. This work shares a similar motivation to ours: broadening policy behavior for better downstream adaptation. We midtrain a PostBC policy on the same standard demonstrations and starting from the same pretrained policy checkpoint (i.e., $\pi_{0.5}$), using an ensemble of 20 members trained on frozen policy features and a noise scale of $\alpha=1$. On the twelve OOD simulation tasks, PostBC achieves a 64.0\% success rate, compared with 62.7\% for standard demonstrations, while still being far below 79.2\% for \methodname{} (Table~\ref{tab:zero-shot-subtask-detailed}). On the eight ID tasks, the success rates are 43.0\%, 42.8\%, and 55.5\% respectively for PostBC, standard demonstrations, and \methodname{}. PostBC thus performs comparably to midtraining on standard demonstrations, but underperforms \methodname{}.

\textbf{Comparison with frontier models.} We also compare with a frontier model, GPT-6.2 Astra, run directly as the robot policy on real-world LEGO disassembly. This comparison tests the motivation for the tasks: the challenge is not primarily one of specifying \emph{what} should happen next, but of \emph{how} to realize that outcome physically. At each step, the model observes the camera images and the current subtask instruction, and outputs an end-effector motion and a gripper command, which we execute after inverse-kinematics and collision checks. The model plans correctly: it holds the assembly with one arm and pulls a brick with the other. However, it takes 88 minutes to take apart and sort a single assembly (Appendix~\ref{app:frontier}). The difficulty lies in low-level execution rather than in the plan, which indicates that this task tests the low-level behavioral coverage.

\begin{figure}[t]
\centering
\vspace{-0.2cm}
\includegraphics[width=0.99\linewidth]{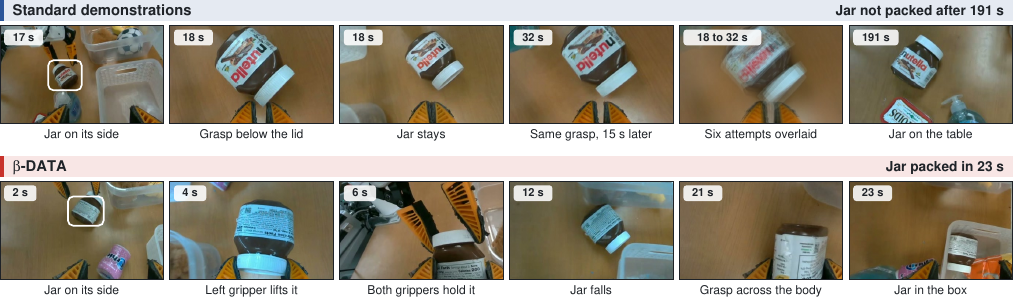}
\vspace{-0.25cm}
\caption{\footnotesize{\textbf{Repeated grasp attempts and recovery in real-world packing.} Both policies are instructed to pack a jar that lies on its side. The policy trained on standard demonstrations (top) repeats the same grasp below the lid, and the jar stays on the table. The policy trained on \methodname{} (bottom) lifts the jar, and when the jar falls, it grasps the jar across its body and packs it.}}
\label{fig:real-xarm-jar}
\vspace{-0.3cm}
\end{figure}
\begin{figure}[t]
\centering
\vspace{-0.2cm}
\includegraphics[width=0.99\linewidth]{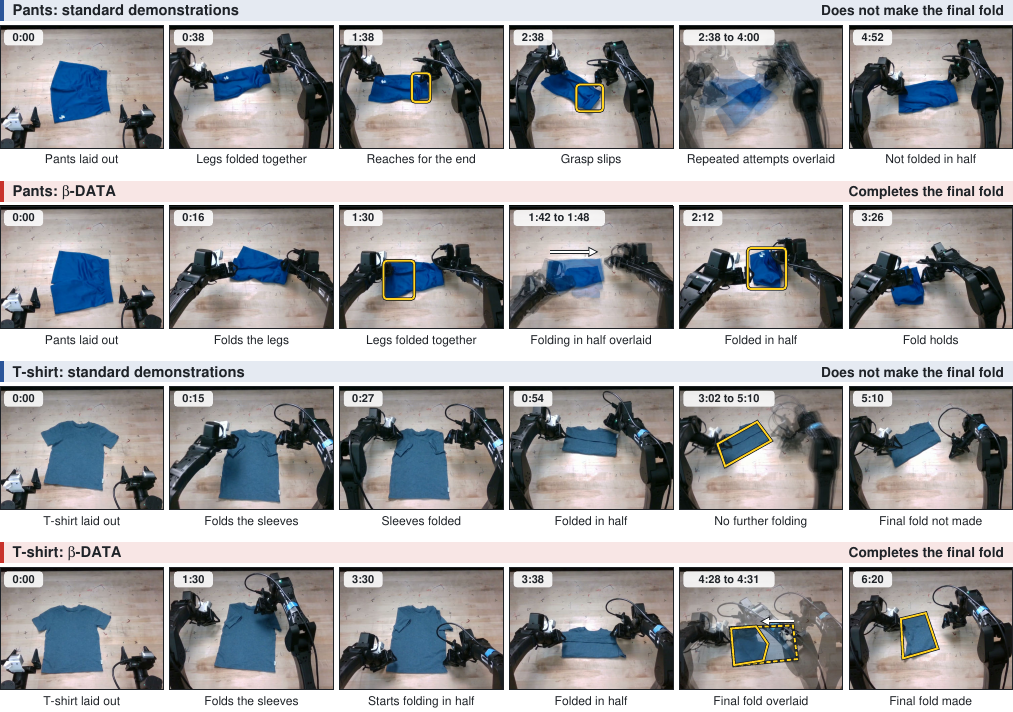}
\vspace{-0.25cm}
\caption{\footnotesize{\textbf{The policy trained on standard demonstrations is usually unable to make the final fold of either garment.} The pants are folded twice: legs together, then in half. The T-shirt is folded three times: sleeves, in half, then in half again, which is the final fold. Yellow outlines mark the garment or the part that the final fold moves. Frames with a time range overlay several frames from that range. \textit{\textbf{Top:}} pants. Both policies fold the legs together. The policy trained on standard demonstrations (first row) loses its grasp on the folded pants, and its repeated attempts do not fold them in half. The policy trained on \methodname{} (second row) folds the pants in half, and the fold holds. \textit{\textbf{Bottom:}} T-shirt, from the same initial layout. Both policies fold the T-shirt in half. The policy trained on standard demonstrations (third row) then stops, and the T-shirt stays unchanged until the rollout ends. The policy trained on \methodname{} (fourth row) makes the final fold.}}
\label{fig:garment-folding-rollouts}
\vspace{-0.3cm}
\end{figure}

\textbf{Qualitative behavior.}
\looseness=-1 In Figures~\ref{fig:sim-task11-recovery} and~\ref{fig:real-xarm-jar}, we illustrate how improving connected behavioral coverage can be crucial even when the desired subtask is already specified. Starting from the same state, both policies are instructed to pack the sugar box into the medium tray. The policy trained on standard demonstrations lifts the sugar box and holds it beside the tray without placing it. The policy trained on \methodname{} places the sugar box, and when the box slides over the tray rim, it regrasps the box, sets it flat in the tray, and completes the task. This example shows that the remaining challenge is not deciding \emph{what} to do, but executing an appropriate low-level realization of the specified subtask, including after a failed attempt. In real-world packing as well (Figure~\ref{fig:real-xarm-jar}), both policies are instructed to pack a jar that lies on its side. The policy trained on standard demonstrations repeatedly attempts the grasp in the same manner, and the jar stays on the table. The policy trained on \methodname{} grasps the jar with one gripper, and when the jar falls back onto the table, it retries, grasps the jar across its body with the other gripper, and packs it. As in Figure~\ref{fig:lego-removal-routes} for LEGO disassembly, this reflects the connected behavioral coverage induced by training on \methodname{}: under the same instruction, the policy packs the jar through three different sequences of grasps (Figure~\ref{fig:real-xarm-jar-routes} in Appendix~\ref{app:real-xarm-rollouts}). We also run PostBC~\citep{wagenmaker2025posteriorbehavioralcloningpretraining} from the state of Figure~\ref{fig:sim-task11-recovery} (Figure~\ref{fig:sim-task11-postbc} in Appendix~\ref{app:sim-packing-filmstrips}). PostBC lifts the sugar box but leaves it on the tray rim, while the policy trained on \methodname{} carries the sugar box into the tray and then packs the remaining objects. In garment folding (Figure~\ref{fig:garment-folding-rollouts}), the policy trained on standard demonstrations does not make the final fold of either garment. On the pants, it loses its grasp and does not fold them in half, while the policy trained on \methodname{} completes this fold. On the T-shirt, it stops after folding the T-shirt in half once, while the policy trained on \methodname{} folds it in half again.

\begin{figure}[t]
\centering
\vspace{-0.2cm}
\includegraphics[width=0.75\linewidth]{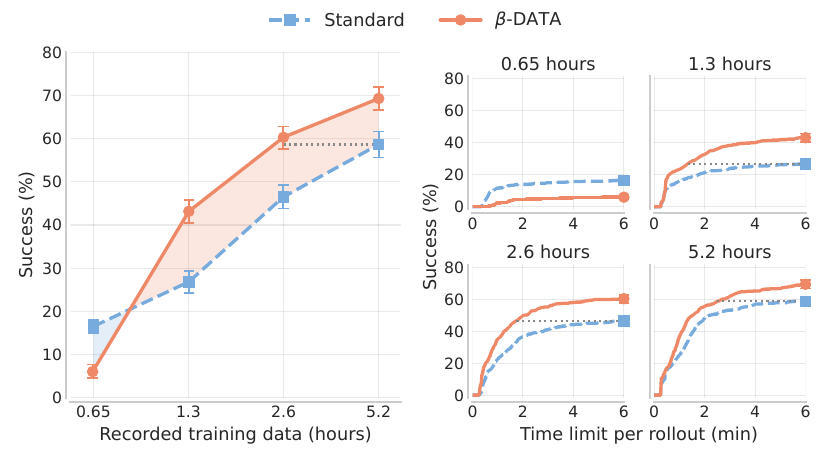}
\vspace{-0.2cm}
\caption{\footnotesize{\textbf{Training data and test-time scaling of \methodname{} in simulation.}
\textit{\textbf{Left:}} zero-shot success as a function of the amount of midtraining data across four matched data budgets. \methodname{} overtakes midtraining on standard demonstrations beyond the smallest budget and achieves comparable performance with roughly half as much data at the larger budgets. \textbf{\textit{Right:}} success as a function of rollout time for policies trained at each data budget. From 1.3 hours onward, \methodname{} outperforms standard demonstrations at every rollout time limit and continues to gain from additional rollout time. Each point is evaluated over 600 trials (12 tasks $\times$ 50); error bars show pointwise 95\% bootstrap intervals over tasks and trials.}}
\label{fig:scaling-budgets}
\vspace{-0.4cm}
\end{figure}

\refstepcounter{resultnum}\label{res:scaling}\textcolor{lightblue}{\textbf{Result~\theresultnum: \methodname{} scales more favorably with both training data and rollout time (i.e., test-time compute).}} Next, we study if the benefits of \methodname{} and its relative gains over standard demonstrations persist as we scale up the amount of data. We train policies on varying amounts of \methodname{} or standard demonstrations and evaluate zero-shot success on the twelve OOD simulation tasks. We use the same batch size of 256 and train for the same number of epochs at every data budget, so the number of training updates increases in proportion to the amount of midtraining data. Training compute therefore scales with the data budget. Figure~\ref{fig:scaling-budgets} (left) shows that while standard demonstrations perform better at the smallest budget (16.5\% vs. 6.0\%), \methodname{} overtakes them thereafter. On the right, we note that the test-time scaling behavior of \methodname{} is substantially better than with standard demonstrations.
Notably, \methodname{} trained on 2.6 hours reaches 60.3\% success, comparable to the 58.7\% achieved by training on \emph{twice} as many hours of standard demonstrations. Here, \methodname{} achieves comparable performance with about \emph{half as much midtraining data}, while continuing to improve as the training data increases.

The same policies also benefit from additional rollout time (i.e., test-time compute), including budgets well beyond typical human completion times. In simulated packing, 86\% of teleoperated demonstrations finish within two minutes and 99\% within four minutes, whereas \methodname{} continues to improve up to six minutes. With 5.2 hours of training data, increasing the rollout limit from two to four to six minutes raises its success rate from 55.5\% to 65.2\% to 69.3\%, compared with 47.8\% to 56.7\% to 58.7\% for the policy trained on standard demonstrations. These gains reflect productive progress during extended rollouts: at the 5.2-hour data budget, 12.0\% of \methodname{} evaluation rollouts still unfinished at four minutes succeed by six minutes, compared with only 4.6\% for standard demonstrations. Thus, \methodname{} benefits from extending rollout time beyond human demonstration lengths and achieves a higher success rate at the longest budget tested.
Alongside these scaling benefits, \methodname{} reaches a given success rate sooner. With 1.3, 2.6, and 5.2 hours of training data, it matches the six-minute success rate of policies trained on the same amount of standard demonstrations in 1.3, 1.7, and 2.5 minutes, respectively, and continues improving afterward. For data budgets of at least 1.3 hours, it outperforms standard demonstrations at every rollout limit. Table~\ref{tab:test-time-budgets} in Appendix~\ref{app:scaling} reports results across all rollout and data budgets.

\vspace{-0.05cm}
\begin{AIbox}{Takeaways: Data and rollout-time scaling}
\begin{itemize}[itemsep=0pt]
 \setlength{\leftskip}{-15pt}
    \item Midtraining on \methodname{} achieves a higher success rate than training on twice as many hours of standard demonstrations, on out-of-distribution tasks.
    \item Extending rollout time from two to six minutes raises \methodname{} success by a larger amount than for standard demonstrations.
\end{itemize}
\end{AIbox}

\begin{wrapfigure}{r}{0.6\linewidth}
\centering
\vspace{-0.2cm}
\includegraphics[width=\linewidth]{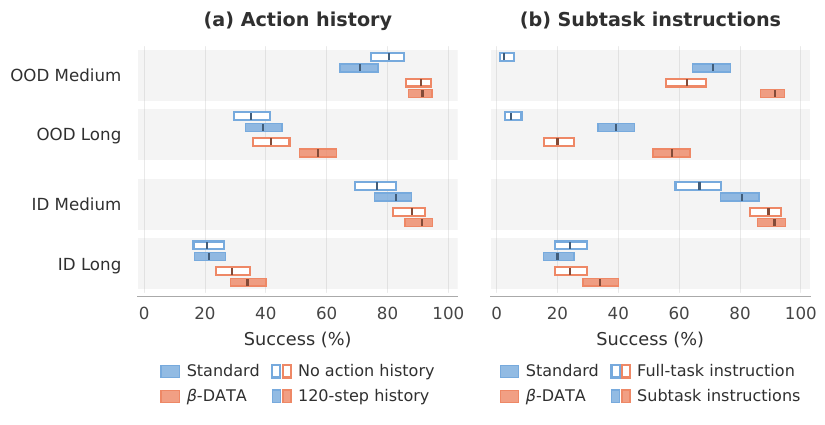}
\vspace{-0.8cm}
\caption{\footnotesize{
\textbf{Ablations of action-history conditioning and language instruction granularity.}
Each ablation reports full-task success averaged equally across tasks in each category.
Standard denotes the policies trained on standard demonstrations. Each box spans the 95\% confidence interval of the success rate, and the dark line inside it marks the mean.
\textbf{(a)} \methodname{} benefits substantially from action-history conditioning on OOD Long tasks, whereas the corresponding gain is much smaller for policies trained on standard demonstrations.
\textbf{(b)} Segment-level outcome annotations and corresponding subtask instructions substantially improve OOD execution for both policies, while \methodname{} retains a clear advantage over midtraining on standard demonstrations.
}}
\label{fig:design-ablations}
\vspace{-0.6cm}
\end{wrapfigure}

\textbf{Ablation 1: Action history \textcolor{magenta}{(D2)} helps sustain behavior over long horizons.} We next ablate the two design choices involved in midtraining on \methodname{}. First, we study whether recent action history helps the policy select how to continue a low-level behavior, beyond what is specified by the current observation and subtask instruction. We compare policies trained on \methodname{} and on standard demonstrations, trained and evaluated either with a 120-step action history or without history, using matched training and evaluation settings otherwise. Figure~\ref{fig:design-ablations}(a) shows that the benefit is largest for \methodname{} on long-horizon tasks. In simulation, on OOD long-horizon tasks, history improves \methodname{} success from 41.6\% to 57.6\%, compared with a 4\% improvement for the policy trained on standard demonstrations, and increases task progress from 62.8\% to 77.6\%. All five OOD Long tasks improve. On ID long-horizon tasks, \methodname{} success increases from 28.8\% to 34.0\%, with task progress increasing from 47.0\% to 58.9\%. In contrast, history has little effect on the easier OOD Short and Medium tasks, where \methodname{} is already near saturation.

These results support the role of action history as context when training on \methodname{}. Long adversarial trajectories in \methodname{} expose the policy to diverse low-level behaviors, and recent actions help distinguish which stage of such a maneuver the policy is currently executing. This is particularly important on long-horizon tasks, where repeated manipulation and rearrangement continually change the constraints faced by the policy. Thus, improving connected behavioral coverage (CBC) of the training dataset alone is not sufficient: the policy must also condition on its recent behavior to select an appropriate continuation from that coverage.

\textbf{Ablation 2: Subtask annotations \textcolor{magenta}{(D1)} improve held-out performance.} We next study the role of the segment-level outcome annotations used during midtraining. We compare the two execution settings from Section~\ref{sec:envs_tasks}: direct execution from a single full-task instruction, and hierarchical execution, where a high-level planner provides the policy with the current subtask. The two settings differ in both the training annotations and the instructions given at execution. The intuition is that a subtask instruction makes the immediate goal explicit, while action history specifies how the current attempt has unfolded. Figure~\ref{fig:design-ablations}(b) shows that providing subtask instructions substantially improves performance on OOD tasks. For \methodname{}, success on OOD Long tasks increases from 20.0\% to 57.6\%, with task progress increasing from 52.6\% to 77.6\%; on OOD Medium tasks, success increases from 62.5\% to 91.5\%. Policies trained on standard demonstrations also benefit, improving from 4.8\% to 39.2\% on OOD Long tasks and from 2.5\% to 71.0\% on OOD Medium tasks. The gains are much smaller and more mixed on ID tasks, suggesting that explicitly specifying the local outcome is most useful when the policy needs to generalize to new tasks.

Finally, we observe that \textbf{\emph{subtask decomposition does not remove the need for increasing connected behavioral coverage}}. Even when policies trained on \methodname{} and on standard demonstrations receive subtask instructions, \methodname{} exceeds the policy trained on standard demonstrations by 20.5\% and 18.4\% on OOD Medium and Long tasks, respectively. However, both policies remain at or below a success rate of 16\% on the hardest OOD rearrangement tasks, showing that specifying \emph{what} outcome to pursue does not by itself determine \emph{how} to realize it. Together with Ablation 1, these results suggest complementary roles for the two design choices: subtask annotations specify the outcome we desire, while action history provides context for selecting an appropriate continuation from the behavioral repertoire of the policy. 

\begin{AIbox}{Takeaways: Subtask instructions and action history}
\begin{itemize}[itemsep=0pt]
 \setlength{\leftskip}{-15pt}
    \item Action-history conditioning is more useful for \methodname{} on longer tasks.
    \item Subtask instructions improve OOD performance for both policies, but \methodname{} retains an advantage over standard demonstrations.
\end{itemize}
\end{AIbox}

\vspace{-0.3cm}
\section{Results: Training on \methodname{} Enables Efficient Online Finetuning}
\label{sec:online-finetuning}
\vspace{-0.2cm}

Next, we study whether training with \methodname{} also improves the resulting policy’s ability to learn from its own experience. Why could this be true? First, broader low-level behavioral coverage should help a policy generate successful experience more easily: a policy that can recover from challenging situations and retry is more likely to succeed. Second, even given the same number of successful rollouts, the content of those rollouts can differ across policies. A policy with broader coverage can succeed from a wider range of states and through more diverse low-level behaviors, potentially providing richer supervision for subsequent on-policy updates\footnote{For analogy, note that large language models (LLMs) that ``think'' for longer and explore better also generalize to harder problems, i.e., exploration enables extrapolation~\citep{setlur2026e3}.}. 
We study these possibilities when finetuning \methodname{} policies via filtered BC. In this procedure, the current policy collects rollouts, admits the successful ones, and is repeatedly updated on a mixture of these admitted successful trajectories and existing offline data.

\refstepcounter{resultnum}\label{res:online-sim} \textcolor{lightblue}{\textbf{Result~\theresultnum: \methodname{} makes useful online experience substantially easier to collect on challenging tasks.}} The largest difference appears on long-horizon tasks, where a successful trajectory may require recovering from local failures while still completing the full task. As shown in Figure~\ref{fig:oodlong-m} (right) and Figure~\ref{fig:online-ood-tasks} in Appendix~\ref{app:online-task-curves}, the total rollout count, averaged across the five OOD Long tasks, is 160 from the \methodname{} initialization, compared with 3,810 from the policy trained on standard demonstrations. Despite this substantially smaller interaction budget, \methodname{} improves from 57.6\% to 73.2\% success, while the policy trained on standard demonstrations improves from 39.2\% to 52.0\%. 
Across these tasks, the performance improvement per rollout is higher for \methodname{} than for standard demonstrations. Thus, the \methodname{} initialization lowers the interaction cost of obtaining experience for online improvement.

\textbf{Fitting a law to online finetuning performance.} A policy that starts with a higher success rate can collect successful trajectories in fewer attempts. We therefore ask how much each policy improves after admitting the same number of successful trajectories. On the held-out long-horizon tasks, we model success after finetuning as a function of the initial success rate $p_0$ and the number of admitted successful trajectories $M$. This follows the distinction between initial model capability and post-training budget in scaling studies of language-model reinforcement learning~\citep{shen2026understanding}. We model $\log\frac{p}{1-p}$ rather than the success rate $p$ itself, so the output of the law can span the entire real line. We use $\log M$ so that each doubling of $M$ adds the same amount, $c\log 2$, to $\log\frac{p_M}{1-p_M}$: doubling the number of successful trajectories from 10 to 20 has the same modeled effect as doubling it from 20 to 40. For each dataset, we fit (Figure~\ref{fig:oodlong-m}):
{\setlength{\abovedisplayskip}{10pt}\setlength{\belowdisplayskip}{4pt}\setlength{\fboxsep}{3pt}%
\begin{align*}
\boxed{\smash[b]{\log\frac{p_M}{1-p_M} \;=\; \underbrace{a + b\,\log\frac{p_0}{1-p_0}\rule[-13pt]{0pt}{13pt}}_{\text{initial success rate}} \;+\; \underbrace{c\,\log{M}\rule[-13pt]{0pt}{13pt}}_{\text{online experience}}}\vphantom{\ln\frac{p_M}{1-p_M}}}\rule[-31pt]{0pt}{31pt}
\end{align*}}

\begin{figure}[t]
\centering
\includegraphics[width=0.45\linewidth]{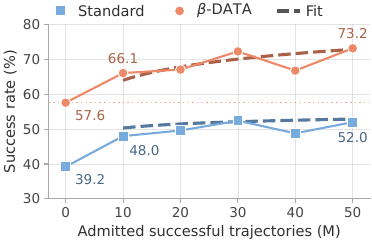}\hspace{0.04\linewidth}%
\includegraphics[width=0.45\linewidth]{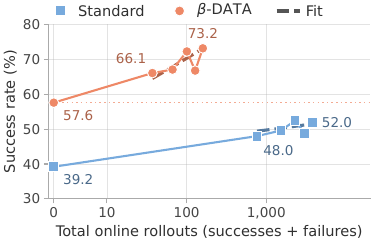}
\vspace{-0.2cm}
\caption{\footnotesize{\textbf{On held-out long-horizon tasks, the policy trained on \methodname{} starts higher and improves more with online experience.} On simulated \textbf{\textit{grocery packing}} tasks, success rate averaged over the five tasks, against the number of admitted successful trajectories $M$ (left) and against the total number of online rollouts, including failures (right). Dashed lines show the law fitted separately to each training dataset. At $M = 50$, the policy trained on standard demonstrations remains below the initial success rate of the policy trained on \methodname{} (dotted line).}}
\label{fig:oodlong-m}
\vspace{-0.3cm}
\end{figure}
\noindent where $p_0$ is the initial success rate on the task and $p_M$ is the success rate after $M = 10, 20, \dots, 50$ admitted successful trajectories. Here, $a$ is the intercept, $b$ captures the effect of the initial success rate, and $c$ the effect of the amount of \emph{\textbf{successful}} online experience. \methodname{} policies start from a higher success rate (57.6\% against 39.2\%) and improve to 73.2\% after fifty admitted successful trajectories, while policies trained on standard demonstrations improve to 52.0\% and remain below the initial success rate of \methodname{} policies. Between $M = 10$ and $M = 50$, the law gives a larger effect of online experience for \methodname{} policies ($c = 0.43$, with a 95\% bootstrap confidence interval over tasks from $0.09$ to $0.74$) than for policies trained on standard demonstrations ($c = 0.20$, from $0.02$ to $0.48$). At a fixed initial success rate, each doubling of $M$ raises $\log\frac{p_M}{1-p_M}$ by $c\log 2$: by $0.30$ for \methodname{} and by $0.14$ for standard demonstrations. This suggests that the utility of each new successful rollout from a \methodname{} policy is higher than that of a policy trained on standard demonstrations.

\begin{wraptable}{r}{0.5\linewidth}
\vspace{-0.3cm}
\centering
\small
\setlength{\tabcolsep}{4pt}
\renewcommand{\arraystretch}{1.12}
\begin{tabular}{@{}lcc@{}}
\toprule
Training data & Success (\%) & Progress (\%) \\
\midrule
Standard demonstrations & $20.0 \rightarrow 20.0$ & $45.0 \rightarrow 68.3$ \\[-2pt]
 & {\scriptsize [8.1, 41.6]} & {\scriptsize [59.4, 77.2]} \\
\addlinespace[3pt]
\textbf{\methodname{} (Ours)} & $60.0 \rightarrow \textbf{85.0}$ & $74.4 \rightarrow \textbf{98.3}$ \\[-2pt]
 & {\scriptsize [64.0, 94.8]} & {\scriptsize [96.7, 100.0]} \\
\bottomrule
\end{tabular}
\vspace{-0.3cm}
\caption{\footnotesize{
\textbf{Real-world bimanual xArm \textbf{\textit{grocery packing}} before and after online finetuning.}
Values report zero-shot $\rightarrow$ post-finetuning performance for each policy, with twenty scored trials at each checkpoint. Brackets give the post-finetuning 95\% Wilson interval for success over the twenty trials and percentile bootstrap interval for progress. Each trial begins from a new initial configuration in the workspace, hence the high variance.
}}
\label{tab:real-online-results}
\vspace{-0.3cm}
\end{wraptable}
Measuring the effect of online experience by the total number of online rollouts, including failures, in place of $M$, with a separate intercept for each task in place of the initial-success term, gives a similar result (Figure~\ref{fig:oodlong-m}, right; $c = 0.46$, from $0.08$ to $0.86$, against $c = 0.20$, from $0.01$ to $0.57$).

\begin{figure}[t]
\centering
\vspace{-0.2cm}
\includegraphics[width=0.99\linewidth]{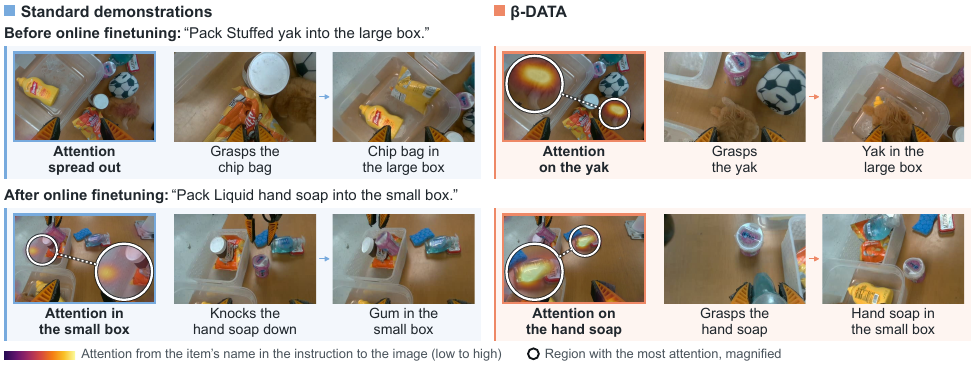}
\vspace{-0.2cm}
\caption{\footnotesize{\textbf{\emph{[Spurious correlations analysis]} In these rollouts, the policy trained on \methodname{} attends to the instructed item before and after online finetuning, while the baseline trained on standard demonstrations attends diffusely or elsewhere.} Real-world bimanual xArm \textbf{\textit{grocery packing}} rollouts. The first frame for each policy shows attention from the name of the item in the instruction to the wrist-camera image, computed on the same frame for both policies; the region receiving the most attention is circled and magnified. The remaining two frames show the rollout of that policy. \textit{\textbf{Top:}} before online finetuning, both policies are instructed to pack the yak. The \methodname{} policy attends tightly to the yak and packs it; the attention of the baseline is diffuse, and it packs the chip bag into the large box. \textit{\textbf{Bottom:}} after online finetuning, both policies are instructed to pack the hand soap, shown after the baseline has knocked it down. The \methodname{} policy attends to the soap and packs it; the baseline attends inside the small box and packs the gum.}}
\label{fig:real-oft-failures}
\vspace{-0.3cm}
\end{figure}
\refstepcounter{resultnum}\label{res:online-real}\textcolor{lightblue}{\textbf{Result~\theresultnum: \methodname{} improves through online finetuning on the bimanual, real-world packing task.}}
We next test whether the policy trained on \methodname{} can improve through successful experience collected on the real-world bimanual xArm packing task. Starting from the midtraining checkpoint, we collect twenty admitted successful trajectories and perform ten gradient updates after each admitted success, for 200 updates in total. Each training batch mixes equal amounts of offline demonstrations and online successful experience. The policy improves from $12/20$ ($60\%$) to $17/20$ ($85\%$) full-task success, while task progress increases from $74.4\%$ to $98.3\%$ (Table~\ref{tab:real-online-results}). Selected rollouts in Figure~\ref{fig:real-xarm-before-after} in Appendix~\ref{app:real-xarm-rollouts} illustrate a change in behavior: the zero-shot policy fails to pack the mustard, while the policy after online finetuning packs it and continues until all items are packed. In contrast, the policy midtrained on standard demonstrations does not improve from $20\%$ success before finetuning (Table~\ref{tab:real-online-results}). It needs 82 rollouts to collect twenty admitted successful trajectories, while the \methodname{} policy only needs 28. 

\textbf{Dynamics of finetuning with \methodname{}.} We next study why \methodname{} enables faster and more effective finetuning on the real robot, compared with the policy trained on standard demonstrations. To ground our understanding of this difference, we examine what the online finetuning process reinforces. As shown in Figure~\ref{fig:real-oft-failures}, policies trained on standard demonstrations appear to rely on some spurious aspects of the scene: before online finetuning, the policy packs the chip bag when it is instructed to pack the yak in 11 of 20 trials. This reliance persists through finetuning, and finetuning reinforces repeated grasping failures of the same kind: after online finetuning, the policy knocks down or drops the hand soap in 12 of 20 trials, and in 6 of these trials it then packs the gum, which the instruction does not name. In contrast, with \methodname{}, reliance on spurious cues present in the initial policy diminishes over time: the policy packs an item that the instruction does not name in 3 of 20 trials before online finetuning and in none after. The attention maps in Figure~\ref{fig:real-oft-failures} show the same difference: the attention from the words that name the item falls on that item for the \methodname{} policy, while for the policy trained on standard demonstrations it is spread out or falls elsewhere, and online finetuning does not fix this issue.

\begin{AIbox}{Takeaways: Online finetuning}
\begin{itemize}[itemsep=0pt]
 \setlength{\leftskip}{-15pt}
 \item \methodname{} policies require substantially fewer rollouts to see an improvement than policies trained on standard demonstrations. Both the initial policy success rate and ``quality'' of successful rollouts contribute to effective online finetuning of the \methodname{} policy.
\end{itemize}
\end{AIbox}

\vspace{-0.2cm}
\section{Related Work}
\vspace{-0.2cm}

\textbf{Data scaling in robot learning.}
The success of foundation models~\citep{kaplan2020scaling, hoffmann2022training, brown2020language} has motivated efforts to scale data, model size, and compute in robot learning. Scaling real-world demonstrations has enabled both dexterous task-specific policies~\citep{zhao2024aloha} and increasingly generalist policies spanning tasks, environments, and embodiments~\citep{o2024open, brohan2023rt}. As robot datasets continue to grow, an important question is therefore not only how much data to collect, but what kind of experience is most productive to scale. Our work focuses on this question at the level of low-level behavior: standard demonstrations, teleoperated or otherwise, provide broad task and state coverage, but may still provide narrow coverage over the different ways a policy can adapt, recover, and continue making progress. In contrast, \methodname{} targets this coverage directly, and our results show that it scales more favorably than standard demonstrations.

\textbf{Data for imitation learning.} Because real-robot data is expensive, several works study how to use a fixed collection budget more effectively. Prior approaches broaden coverage across objects, environments, initial poses, or combinations of these factors~\citep{lin2024data,gao2024efficient,zha2025guiding,xiao2026towards}. However, the resulting trajectories still largely consist of successful expert execution and do not necessarily improve low-level behavioral coverage. Play data~\citep{lynch2019play} provides broader experience through task-agnostic interaction, but does not deliberately interrupt task progress to elicit alternative low-level behaviors. A complementary line of work collects corrections or recoveries during policy execution~\citep{wu2025robocopilot,hu2026rac}. While this exposes diverse behaviors from challenging states, the resulting data distribution remains tied to the failures of the base policy, making it harder to collect reusable training data independently of the downstream learner. Injecting noise into the expert's actions during data collection, or predicting action chunks, avoids the exponential compounding of imitation errors~\citep{zhang2025action}. Adversarial data collection~\citep{huang2025adversarial} instead perturbs the environment or task during human teleoperation while the operator continues toward completion, naturally eliciting retries, strategy changes, and recovery without learner rollouts. We build on this idea in \methodname{} and use these trajectories as midtraining data for pretrained policies. Our contribution is to study how the resulting behavioral coverage affects generalization, scaling, and online improvement. To the best of our knowledge, we are also the first to scale up adversarial data collection to complex multi-stage, bimanual tasks.

\textbf{Exploratory behaviors in robot policies.}
Broad behavioral coverage can be useful in two related ways.
First, it can allow a policy to continue attempting actions in order to identify a successful realization of a task \emph{within} a single rollout.
Analogously, language and vision-language models can revise intermediate solutions through reasoning and interaction~\citep{wei2022chain,yao2022react,zawalski2024robotic,guo2025deepseek}, while recurrent policies can adapt their behavior based on previous observations and actions~\citep{duan2016rl}.
In robotics, this motivates policies that can retry, recover, or change low-level strategy rather than committing to a single execution attempt or strategy. Second, a policy's behavioral coverage determines the experience it can generate for subsequent learning.
Reinforcement-learning methods encourage exploration through action noise~\citep{fujimoto2018addressing}, entropy maximization~\citep{pmlr-v80-haarnoja18b}, or intrinsic rewards~\citep{pathak2017curiosity,burda2018exploration}, while offline approaches learn behavioral priors, skills, or exploratory policies that improve subsequent data collection~\citep{singh2020parrot,pertsch2021accelerating,pmlr-v267-wagenmaker25a}.
Robot policies can also improve from their own experience~\citep{bousmalis2023robocat}. Recent work finetunes pretrained vision-language-action policies with online reinforcement learning~\citep{guo2025improving,chen2025conrft}, or lets the robot practice downstream tasks autonomously with a reward function and a success detector obtained from the pretrained model~\citep{ghasemipour2025self}. These methods study how to learn from online experience given a pretrained policy. Our work is complementary: we study how the midtraining demonstrations shape the experience that a policy collects for online finetuning.
Recent work also broadens pretrained robot-policy behavior by modeling uncertainty in demonstrator actions~\citep{wagenmaker2025posteriorbehavioralcloningpretraining} or using timestep-modulated pretraining~\citep{hong2026tmrl}. We also compare our approach with PostBC~\citep{wagenmaker2025posteriorbehavioralcloningpretraining} in Section~\ref{sec:zero-shot-tasks}, and find that PostBC broadens the action distribution without improving behavioral coverage meaningfully. In contrast, we aim to build exploratory behavior into a pretrained robot policy by midtraining it on \methodname{}.
The resulting policy can use alternative low-level behaviors to sustain progress on unfamiliar configurations and generate experience for subsequent finetuning.

\textbf{Frontier models for robot control.} Recent work studies frontier models for both planning and direct robot control. Gemini Robotics is a vision-language-action model trained for robot control~\citep{geminiroboticsteam2025}. Other work runs a general-purpose frontier model directly as the robot policy. GPT-6 Astra produces end-effector targets for simulated manipulation tasks without task-specific finetuning~\citep{zhang2026unexpected}, and controls the robot on selected long-horizon household tasks in simulation through observation and motion tools, starting from its own practice on training instances~\citep{larchenko2026one}.
These approaches remain limited in low-level execution. Code-as-policies methods can only use the behaviors that their perception modules can describe and their control primitives provide~\citep{liang2023code,huang2023voxposer}. When a frontier model produces the actions directly, it generalizes on tasks that require semantic understanding, but performs poorly on tasks that require precision, dynamic control, or bimanual coordination~\citep{zhang2026unexpected}, and a household episode takes about 17 times longer than real time~\citep{larchenko2026one}. \methodname{} targets this low-level execution: we study how midtraining demonstrations affect low-level behavior once a subtask instruction is given.

\vspace{-0.2cm}
\section{Discussion and Perspectives on Future Work}
\label{sec:conclusion}
\vspace{-0.2cm}

In this paper, we studied whether broadening coverage over low-level behaviors improves robot policies and their ability to learn from autonomous interaction (``post-training''). Our tasks remain challenging even when a frontier LLM-level high-level planner specifies what to do next, and our results show that deliberately collecting this form of midtraining data improves policy performance and adaptation. Our results also suggest benefits to feature learning arising from broader and more intricate state-action coverage offered by \methodname{}. More rigorously understanding how broader behavioral coverage affects the features a policy learns, and how these features contribute to downstream performance and improvement, is an important direction for future work. It remains unclear whether comparable benefits would emerge simply by scaling naturally collected demonstrations in a ``grokking'' fashion, or whether deliberate changes to data collection are needed.

These behaviors also raise questions about how to enable more autonomous improvement. A policy trained to recover from perturbed states may be able to continue from the states left by its own failed attempts or execute behaviors needed to reset the environment. Frontier models could also help coordinate resets using learned low-level policies. Understanding when to continue pursuing a goal, when to reset, and how to carry out these resets could make autonomous practice more effective. There is also an opportunity to combine \methodname{} with intervention-based collection. Human interventions provide targeted corrections for the states visited and mistakes made by a particular policy, while \methodname{} can be collected independently of that policy. How to combine this reusable midtraining data with policy-specific corrections and autonomous experience remains an open question.

Finally, we would like to understand how \methodname{} scales across more tasks, embodiments, and adversaries. Our protocol requires only that perturbations be reversible by the operator; their form is otherwise unrestricted, so the collected data can vary substantially across adversaries. In our experiments, data was collected by 10 different  people. Despite this variation, we observe consistent scaling trends in our experiments. This motivates a question analogous to those studied in language modeling: can heterogeneous human-generated data yield predictable improvements as collection scales? Testing whether the benefits of \methodname{} persist at larger scales is a direct next step. Automating the adversary, or standardizing aspects of the protocol without restricting behavioral diversity, could make this collection easier. Understanding the trade-offs between standardization and embracing diversity and heterogeneity is another interesting direction for future work.

\vspace{-0.2cm}
\section*{Acknowledgments}
\vspace{-0.2cm}
We thank Saksham Singh, Sreyas Venkataraman, Anthony Liang, Paarth Shah, Masha Itkina, and Patrick Miller for feedback and discussions. We thank Limin Wei, Naveen Enock, Jeffrey Yu, Niharika Pant, and David Shustin for their help with robot setups and data collection. We thank members of the CMU AIRe and RCHI labs, and UW WEIRD lab for their support. This work was primarily funded by the Toyota Research Institute U2.0 and U3.0 programs. We thank the Babel computing cluster at Carnegie Mellon University, the TPU research cloud (TRC) program of Google Cloud, the Delta AI cluster at NCSA, and NVIDIA for compute support associated with different parts of the project. We also thank the CMU FLAME center for providing compute resources on the Orchard cluster.

\vspace{-0.2cm}
\section*{Author Contributions}
\vspace{-0.1cm}
ZH initially proposed the idea of studying exploration via data collection protocols, and K formulated the data collection protocol to be used in practice and scaled it up with input from ZH. K designed and built the simulation environments with guidance from ZH and others, and set up the real-world packing task. ZH and RW designed the LEGO task in the real world. YZ, SY, KT, and VT from UW owned and designed the garment folding task with inputs from K and ZH. K and ZH designed different parts and versions of the codebase, designed and debugged policy training experiments in different domains of this paper. ZH proposed and implemented the action history component of the VLA policy. K wrote the formal definition of connected behavioral coverage and its measurement with inputs from AK and ZH. K set up the online finetuning experiments in simulation and ran real-world packing online finetuning. Both K and ZH performed several analyses with inputs from AK, many of which did not make it into the final paper. YL and RW helped with data collection, robot policy evaluations, and online finetuning experiments on the real robots at CMU. Both K and ZH designed, performed, and analyzed the experiments and wrote the paper. K, ZH, and AK worked on writing the manuscript and polishing the overall narrative and story. K ran the comparison to frontier models. DS provided feedback on the initial idea development, AG supervised the UW part of the project, and AG, ZE, AO, and KA gave regular feedback on the underlying research process and the manuscript. All authors contributed to ideas and feedback at various stages of the project, helped with editing, and contributed to the overall project.

\bibliography{main}
\bibliographystyle{plainnat}

\clearpage
\appendix

\part*{Appendices}
\makeatletter
\setlength{\@fptop}{0pt}
\setlength{\@fpbot}{0pt plus 1fil}
\setlength{\@fpsep}{14pt plus 2pt}
\makeatother
\renewcommand{\topfraction}{0.9}
\renewcommand{\bottomfraction}{0.8}
\renewcommand{\textfraction}{0.07}
\renewcommand{\floatpagefraction}{0.92}
\setcounter{topnumber}{3}
\setcounter{bottomnumber}{2}
\setcounter{totalnumber}{4}

\section{Simulation task descriptions}
\label{app:packing-tasks}

This appendix describes the twenty simulated packing tasks that we evaluate in Sections~\ref{sec:zero-shot-tasks} and~\ref{sec:online-finetuning}, and defines the short, medium, and long categories used in Table~\ref{tab:zero-shot-subtask}. The twenty evaluation tasks consist of eight in-distribution (ID) training tasks and twelve held-out out-of-distribution (OOD) tasks.

\textbf{Task categories.} We group tasks by the object-level changes required from their initial configurations. Short-horizon tasks require one or two object-level outcomes, such as a placement or the removal of an obstruction. Medium-horizon tasks require three to five direct placements. Long-horizon tasks require either three or more changes, at least one of which removes or transfers existing tray contents, or the placement of eight to ten objects. The OOD split contains three short-horizon, four medium-horizon and five long-horizon tasks. The ID split contains three medium-horizon and five long-horizon tasks.

\textbf{Subtask counts.} Table~\ref{tab:packing-task-inventory} counts one subtask for each required placement, transfer or obstruction removal. An object that is already in its required location adds no subtask, and moving an object from one tray to another counts as one subtask.

\begin{table}[htbp]
\centering
\footnotesize
\setlength{\tabcolsep}{4pt}
\renewcommand{\arraystretch}{1.12}
\begin{tabularx}{\linewidth}{@{}ll>{\raggedright\arraybackslash}Xr@{}}
\toprule
Split & Task & Full goal (summary) & Subtasks \\
\midrule
\multicolumn{4}{@{}l}{\textbf{Short}} \\[3pt]
OOD & Task~1 & Move the gelatin box from medium to small, then pack the foam brick into small. & 2 \\[2pt]
OOD & Task~2 & Pack the gelatin box and foam brick into the medium tray alongside the sugar box already there. & 2 \\[2pt]
OOD & Task~7 & Remove the soup can obstructing the small tray while retaining the gelatin box and foam brick there. & 1 \\
\midrule
\multicolumn{4}{@{}l}{\textbf{Medium}} \\[3pt]
OOD & Task~9 & Pack the gelatin box and foam brick into medium; pack the sugar box and tuna can into large. & 4 \\[2pt]
OOD & Task~10 & Pack the soup can into small, the sugar box into medium, and the cracker box and gelatin box into large. & 4 \\[2pt]
OOD & Task~11 & Pack the foam brick and sugar box into medium; pack the soup can, tuna can, and gelatin box into large. & 5 \\[2pt]
OOD & Task~12 & Pack the gelatin box into small, the foam brick and sugar box into medium, and the soup can and cracker box into large. & 5 \\
\addlinespace[5pt]
ID & Task~1 & Pack the cracker box, soup can, gelatin box, and foam brick into the large tray. & 4 \\[2pt]
ID & Task~3 & Pack the cracker box and soup can into the medium tray; the sugar box and foam brick into the large tray. & 4 \\[2pt]
ID & Task~4 & Pack the cracker box, soup can, and gelatin box into the large tray; leave the foam brick and tuna can unpacked. & 3 \\
\midrule
\multicolumn{4}{@{}l}{\textbf{Long}} \\[3pt]
OOD & Task~3 & Remove the cracker box from medium, then pack the sugar box and foam brick there; keep the gelatin box in small. & 3 \\[2pt]
OOD & Task~4 & Move the gelatin box and foam brick from medium to small, then pack the sugar box into medium; retain the soup can there. & 3 \\[2pt]
OOD & Task~5 & Five objects start in large: keep the cracker box there, move the sugar box and soup can to medium, and the gelatin box and foam brick to small. & 4 \\[2pt]
OOD & Task~6 & Move the soup can from small to medium and the gelatin box and foam brick from medium to small. Pack the sugar box into medium. & 4 \\[2pt]
OOD & Task~8 & Redistribute five misplaced objects: the cracker box to large, the sugar box and soup can to medium, and the gelatin box and foam brick to small. & 5 \\
\addlinespace[5pt]
ID & Task~2 & Clear three obstructions from the medium tray, then pack the sugar box and gelatin box there. & 5 \\[2pt]
ID & Task~5 & Route six specified objects to the large tray; place the mug and can opener in the medium tray. & 8 \\[2pt]
ID & Task~6 & Sort nine objects by size and type: large items to large, cans and the canister to medium, small flat items to small. & 9 \\[2pt]
ID & Task~7 & Sort ten objects: food and food containers to large, the mug to medium, and the can opener and foam brick to small. & 10 \\[2pt]
ID & Task~8 & Distribute ten objects by explicit assignment across three trays: four to large, three to medium, and three to small. & 10 \\
\bottomrule
\end{tabularx}
\caption{\textbf{Simulation tasks.} Tasks are grouped by the horizon categories defined above, and each group shows its OOD tasks first and then its ID tasks.}
\label{tab:packing-task-inventory}
\end{table}

\section{Robot setups}
\label{app:robot-setup}
This appendix describes the three bimanual robot platforms of Section~\ref{sec:envs_tasks}: the xArm setup for real-world grocery packing, the YAM setup for LEGO disassembly and sorting, and the Raiden YAM setup for garment~folding.

\textbf{xArm setup.}
The platform uses two 7-DoF UFactory xArm~7 arms with xArm grippers fitted with scaled-down soft grippers.
An overhead ZED Mini stereo camera and two wrist-mounted Intel RealSense D405 cameras provide RGB observations at 60\,Hz.
Operators teleoperate the arms with Meta Quest 3 VR controllers, following the setup of RaC~\citep{hu2026rac}: controller motions are mapped to end-effector motions, and we read the controller poses with the \texttt{oculus\_reader} package.\footnote{\url{https://github.com/rail-berkeley/oculus_reader}}

\textbf{YAM setup.}
The platform uses two 6-DoF I2RT YAM arms with two-finger grippers.
Three global-shutter USB RGB cameras, one overhead and one on each wrist, capture $1920\times1080$ images at 60\,Hz with fixed exposure and white balance.
Robot states streamed at 200\,Hz are interpolated to the shared timestamp of each camera frame set.
The state and absolute action targets each contain 12 arm joint positions and two gripper positions.
Two Dynamixel-based GELLO leader arms are read at 200\,Hz, with their latest positions applied and recorded at 60\,Hz.
On the xArm and YAM setups, operators mark subtask boundaries with a foot pedal and label each episode as a success or a failure. 
We label the trajectories after collection to form language-labeled subtask training episodes and resize images to $224\times224$.

\textbf{Raiden YAM setup.}
The Raiden~\citep{raiden2026} platform consists of two 6-DoF I2RT YAM arms equipped with parallel grippers. RGB observations are streamed at 30\,Hz using one overhead Intel RealSense D435 camera and two wrist-mounted Intel RealSense D405 cameras. A second pair of 6-DoF YAM arms serves as the leader devices for teleoperation, continuously sending their latest poses as control commands to the follower arms. During data collection, operators mark subtask boundaries using buttons integrated into the leader handles. After collection, each trajectory is segmented and annotated with language labels to form subtask-level training episodes. All RGB observations are resized to $224 \times 224$ before training.

\section{Task definitions and progress scores}
\label{app:task-progress}

This appendix defines the success rate and the task progress reported in Table~\ref{tab:zero-shot-subtask} (packing ratio in Table~\ref{tab:zero-shot-subtask-detailed}). A trial succeeds only when it completes all subtask requirements. For task progress in packing and LEGO disassembly, we divide the score of each trial by its maximum possible score and multiply by 100. The reported progress is the mean of these percentages across evaluation trials.

\subsection{Simulated grocery packing}
\label{app:sim-packing-progress}

Two simulated xArm~7 arms pack grocery objects in the
containers specified by the task instruction. Some tasks
also require removing obstructions or transferring previously
packed objects. Each target object receives one point for
being packed into its assigned container while following
the instruction for that object, and zero otherwise.
For a task with $N$ target objects, the maximum score is $N$. 
The task goals and horizon categories are listed in
Appendix~\ref{app:packing-tasks}.

\subsection{Real-world grocery packing (xArm)}

Two xArm~7 arms pack nine grocery items into two containers.
An item receives one point only when it is packed into its assigned container while following its instruction. The maximum score is therefore nine points, and each correctly packed item contributes one ninth to the progress of a trial. Full-task success requires packing all nine items correctly.

\subsection{LEGO disassembly and sorting (YAM)}

Two YAM arms disassemble interlocking LEGO structures and
sort the individual bricks into their assigned trays.
We evaluate two configurations: \textbf{YAM interlock}
contains one assembly of six bricks, while \textbf{YAM double}
contains two assemblies with nine bricks in total.

Each brick receives half a point for being separated from
the assembly and another half for placement in its correct
tray. The maximum scores are therefore six points for
YAM interlock and nine points for YAM double. Separating all bricks without sorting them gives 50\% progress. Separating and correctly sorting all bricks gives 100\% progress and completes the task.

\subsection{Garment folding (Raiden YAM)}

The robot folds pants that are initially laid flat. It first folds the legs together, and then folds the garment in half. A trial is successful only when both folds are completed.

Progress is the average of the success rates of the two folds, multiplied by 100. We compute the success rate of the second fold over the trials in which the first fold succeeds.

For T-shirt folding, the robot folds both sleeves, folds the garment in half, and folds it in half again. A trial is successful only when all three folds are completed. Progress is the average of the success rates of the three folds, multiplied by 100. We compute the success rate of each fold over the trials in which the previous fold succeeds. We evaluate each policy in twenty trials on each garment.

\section{Per-task zero-shot simulation results}
\label{app:zero-shot-simulation}

Table~\ref{tab:zero-shot-subtask-detailed} reports the per-task results that the simulation part of Table~\ref{tab:zero-shot-subtask} summarizes. Standard denotes the policy trained on standard demonstrations. The packing ratio is the fraction of objects correctly placed out of the total number of objects, averaged across trials. For a task with $c$ successes in 50 trials, pass@$k$ is computed as $1-\binom{50-c}{k}/\binom{50}{k}$, with $\binom{50-c}{k}=0$ when $50-c<k$. Pass@$k$ measures success within $k$ attempts, with resets between attempts. 

\begin{table}[htbp]
\centering
\small
\setlength{\tabcolsep}{3.5pt}
\renewcommand{\arraystretch}{1.08}
\begin{tabular*}{\linewidth}{@{\extracolsep{\fill}}lrrrr@{\hspace{2em}}rrrr@{}}
\toprule
& \multicolumn{4}{c}{\textbf{Standard}} & \multicolumn{4}{c}{\textbf{\methodname{}}} \\
\cmidrule(lr){2-5}\cmidrule(lr){6-9}
Task & Success & \shortstack{Packing\\ratio} & Pass@5 & Pass@10 & Success ($\Delta$) & \shortstack{Packing\\ratio} & Pass@5 & Pass@10 \\
\midrule
\multicolumn{9}{@{}l}{\textbf{Out-of-distribution (OOD)}} \\[2pt]
\multicolumn{9}{@{}l}{\textit{Short}} \\[1pt]
Task 1 & \textbf{100} & 100.0 & 100 & 100 & \makebox[2.2em][r]{\textbf{100}}\hspace{0.35em}\makebox[3.1em][r]{{\scriptsize $(0)$}} & 100.0 & 100 & 100 \\
Task 2 & \textbf{98} & 99.3 & 100 & 100 & \makebox[2.2em][r]{\textbf{98}}\hspace{0.35em}\makebox[3.1em][r]{{\scriptsize $(0)$}} & 99.3 & 100 & 100 \\
Task 7 & 74 & 87.0 & 100 & 100 & \makebox[2.2em][r]{\textbf{98}}\hspace{0.35em}\makebox[3.1em][r]{{\scriptsize $(+24)$}} & 99.0 & 100 & 100 \\
\addlinespace[4pt]
\multicolumn{9}{@{}l}{\textit{Medium}} \\[1pt]
Task 9 & 92 & 96.0 & 100 & 100 & \makebox[2.2em][r]{\textbf{100}}\hspace{0.35em}\makebox[3.1em][r]{{\scriptsize $(+8)$}} & 100.0 & 100 & 100 \\
Task 10 & 64 & 83.5 & 100 & 100 & \makebox[2.2em][r]{\textbf{86}}\hspace{0.35em}\makebox[3.1em][r]{{\scriptsize $(+22)$}} & 90.5 & 100 & 100 \\
Task 11 & 50 & 66.4 & 97 & 100 & \makebox[2.2em][r]{\textbf{90}}\hspace{0.35em}\makebox[3.1em][r]{{\scriptsize $(+40)$}} & 92.8 & 100 & 100 \\
Task 12 & 78 & 93.6 & 100 & 100 & \makebox[2.2em][r]{\textbf{90}}\hspace{0.35em}\makebox[3.1em][r]{{\scriptsize $(+12)$}} & 95.6 & 100 & 100 \\
\addlinespace[4pt]
\multicolumn{9}{@{}l}{\textit{Long}} \\[1pt]
Task 3 & 70 & 86.7 & 100 & 100 & \makebox[2.2em][r]{\textbf{90}}\hspace{0.35em}\makebox[3.1em][r]{{\scriptsize $(+20)$}} & 95.3 & 100 & 100 \\
Task 4 & 64 & 87.5 & 100 & 100 & \makebox[2.2em][r]{\textbf{86}}\hspace{0.35em}\makebox[3.1em][r]{{\scriptsize $(+22)$}} & 92.5 & 100 & 100 \\
Task 5 & 4 & 26.8 & 19 & 36 & \makebox[2.2em][r]{\textbf{16}}\hspace{0.35em}\makebox[3.1em][r]{{\scriptsize $(+12)$}} & 55.2 & 60 & 86 \\
Task 6 & 56 & 81.5 & 99 & 100 & \makebox[2.2em][r]{\textbf{86}}\hspace{0.35em}\makebox[3.1em][r]{{\scriptsize $(+30)$}} & 90.0 & 100 & 100 \\
Task 8 & 2 & 37.6 & 10 & 20 & \makebox[2.2em][r]{\textbf{10}}\hspace{0.35em}\makebox[3.1em][r]{{\scriptsize $(+8)$}} & 55.2 & 42 & 69 \\
\addlinespace[3pt]
OOD mean & 62.7 & 78.8 & 85 & 88 & \makebox[2.2em][r]{\textbf{79.2}}\hspace{0.35em}\makebox[3.1em][r]{{\scriptsize $(+16.5)$}} & 88.8 & 92 & 96 \\
\midrule
\multicolumn{9}{@{}l}{\textbf{In-distribution (ID)}} \\[2pt]
\multicolumn{9}{@{}l}{\textit{Medium}} \\[1pt]
Task 1 & 76 & 85.5 & 100 & 100 & \makebox[2.2em][r]{\textbf{96}}\hspace{0.35em}\makebox[3.1em][r]{{\scriptsize $(+20)$}} & 96.0 & 100 & 100 \\
Task 3 & 74 & 83.0 & 100 & 100 & \makebox[2.2em][r]{\textbf{82}}\hspace{0.35em}\makebox[3.1em][r]{{\scriptsize $(+8)$}} & 87.5 & 100 & 100 \\
Task 4 & 92 & 96.7 & 100 & 100 & \makebox[2.2em][r]{\textbf{96}}\hspace{0.35em}\makebox[3.1em][r]{{\scriptsize $(+4)$}} & 99.3 & 100 & 100 \\
\addlinespace[4pt]
\multicolumn{9}{@{}l}{\textit{Long}} \\[1pt]
Task 2 & 90 & 92.0 & 100 & 100 & \makebox[2.2em][r]{\textbf{96}}\hspace{0.35em}\makebox[3.1em][r]{{\scriptsize $(+6)$}} & 98.0 & 100 & 100 \\
Task 5 & 10 & 36.8 & 42 & 69 & \makebox[2.2em][r]{\textbf{42}}\hspace{0.35em}\makebox[3.1em][r]{{\scriptsize $(+32)$}} & 65.5 & 94 & 100 \\
Task 6 & 0 & 24.2 & 0 & 0 & \makebox[2.2em][r]{\textbf{18}}\hspace{0.35em}\makebox[3.1em][r]{{\scriptsize $(+18)$}} & 48.7 & 65 & 89 \\
Task 7 & 0 & 30.6 & 0 & 0 & \makebox[2.2em][r]{\textbf{10}}\hspace{0.35em}\makebox[3.1em][r]{{\scriptsize $(+10)$}} & 45.2 & 42 & 69 \\
Task 8 & 0 & 23.4 & 0 & 0 & \makebox[2.2em][r]{\textbf{4}}\hspace{0.35em}\makebox[3.1em][r]{{\scriptsize $(+4)$}} & 37.2 & 19 & 36 \\
\addlinespace[3pt]
ID mean & 42.8 & 59.0 & 55 & 59 & \makebox[2.2em][r]{\textbf{55.5}}\hspace{0.35em}\makebox[3.1em][r]{{\scriptsize $(+12.8)$}} & 72.2 & 78 & 87 \\
\midrule
Overall & 54.7 & 70.9 & 73 & 76 & \makebox[2.2em][r]{\textbf{69.7}}\hspace{0.35em}\makebox[3.1em][r]{{\scriptsize $(+15.0)$}} & 82.1 & 86 & 92 \\
\bottomrule
\end{tabular*}
\caption{\textbf{Zero-shot performance with subtask instructions.} A high-level planner provides each policy with subtask instructions. Within each split, tasks follow the structural categories in Appendix~\ref{app:packing-tasks}. Each task uses 50 trials per policy and a limit of 21,600 steps, which is six minutes at 60\,Hz. All metrics are percentages, and higher is better. Bold marks the higher observed success rate in each row, with both values bold on ties. Packing ratio measures the fraction of objects correctly placed out of the total number of objects. Parentheses show the difference in success rate from Standard. Pass@$k$ estimates success within $k$ attempts, as defined in the text. We weight tasks equally in the split and overall averages. Each policy has 600 OOD trials, 400 ID trials, and 1,000 trials overall. Each policy has one training run.}
\label{tab:zero-shot-subtask-detailed}
\end{table}

\section{Policy training settings}
\label{app:training-settings}

Table~\ref{tab:training-settings} gives the settings we use to midtrain the simulated packing policies in Section~\ref{sec:method} and to finetune them online in Section~\ref{sec:online-finetuning}.

\begin{table}[htbp]
\centering
\footnotesize
\setlength{\tabcolsep}{4pt}
\renewcommand{\arraystretch}{1.12}
\begin{tabularx}{\linewidth}{@{}>{\raggedright\arraybackslash}p{0.19\linewidth}>{\raggedright\arraybackslash}X>{\raggedright\arraybackslash}X@{}}
\toprule
Setting & Midtraining & Online finetuning \\
\midrule
Initialization & Released $\pi_{0.5}$ base checkpoint & Final midtraining checkpoint from the same dataset \\
\addlinespace[2pt]
Trainable modules & Image encoder; action expert; rank-16 backbone adapters; history network and projections & Rank-32 action-expert adapters; rank-16 backbone adapters; history network and projections \\
Frozen weights & Base language-backbone weights & Image encoder and base weights of the language backbone and action expert \\
Trainable parameters & 874M of 3.38B & 53.6M: 22.1M action-expert adapters, 27.9M backbone adapters, and 3.6M history network and projections \\
\addlinespace[2pt]
Optimizer & AdamW: $\beta_1=0.9$, $\beta_2=0.95$, $\epsilon=10^{-8}$, weight decay $10^{-10}$; gradient-norm clipping at 1 & Same as midtraining \\
Learning rate & Linear warmup over the first 5\% of training to $10^{-4}$; cosine decay to $10^{-5}$ & Linear warmup over ten updates to $3\times10^{-5}$; then constant \\
Training budget & 10 epochs over the training dataset & One successful rollout and ten gradient updates per round, for 50 rounds and 500 updates in total \\
Batch & 256 examples from the training dataset & 128 offline and 128 online examples \\
Weight averaging & None & Per-round exponential moving average; decay $\min\{0.95,(k+1)/(k+10)\}$ after $k\geq1$ rounds (0.85 at round 50) \\
\addlinespace[2pt]
Actions and history & 60-action chunks at 60\,Hz; 14 action dimensions; 120-action history relative to the current state (gripper command absolute), encoded as four tokens; history noise (std.\ 0.015) and dropout (0.2) during training & Same as midtraining \\
Normalization & Per-coordinate 1st and 99th percentiles of the dataset mapped to $-1$ and $1$ & Fixed midtraining statistics \\
\addlinespace[2pt]
Online replay & Not applicable & Equal trajectory weights within the fresh and earlier groups; uniform sampling of eligible timesteps within each trajectory \\
Eligible online frames & Not applicable & At least one state coordinate changes by $10^{-3}$ or more; use all frames if fewer than 10\% qualify \\

\bottomrule
\end{tabularx}
\caption{\textbf{Training settings for simulated packing.} Standard and \methodname{} share these settings, with separate datasets and normalization statistics. The LoRA adapters act on the attention and feed-forward weights. In each round of online finetuning, we add one successful rollout to the training data (Appendix~\ref{app:online-task-curves}).}
\label{tab:training-settings}
\end{table}

\subsection{Demonstration datasets}
\label{app:demo-datasets}

We collect teleoperated demonstrations in four task domains. In each domain, we compare standard demonstrations with \methodname{} of about the same number of hours (Table~\ref{tab:demo-datasets}). Demonstrations are recorded at 60\,Hz, except garment demonstrations, which are recorded at 30\,Hz. All domains use 60-action chunks and a 120-action history. For simulated packing, each policy is initialized from $\pi_{0.5}$ and midtrained for 50,000 updates with batch size 256, which is about ten passes through its dataset. The data scaling study in Appendix~\ref{app:scaling} uses matched data budgets of 0.65, 1.3, 2.6, and 5.2 hours.

\begin{table}[h]
\centering
\small
\begin{tabular}{@{}lrr@{}}
\toprule
Task & Standard demonstrations (hours) & \methodname{} (hours) \\
\midrule
Simulated grocery packing & 6.1 & 5.9 \\
Real-world grocery packing (xArm) & 4.4 & 4.9 \\
LEGO disassembly and sorting (YAM) & 4.0 & 4.0 \\
Pants folding (Raiden YAM) & 8.1 & 7.9 \\
T-shirt folding (Raiden YAM) & 4.2 & 4.6 \\
\bottomrule
\end{tabular}
\caption{\textbf{Demonstration datasets.} Hours of standard demonstrations and of \methodname{} for each task.}
\label{tab:demo-datasets}
\end{table}

\section{Data and rollout-time scaling details}
\label{app:scaling}

This appendix gives the rollout-time results shown on the right of Figure~\ref{fig:scaling-budgets} (Result~\ref{res:scaling} in Section~\ref{sec:zero-shot-tasks}). We train policies on four data budgets of standard demonstrations (Standard) or \methodname{} and evaluate zero-shot success on the twelve OOD simulation tasks. As in Section~\ref{sec:zero-shot-tasks}, we use the same batch size of 256 and the same number of training epochs at every data budget.

\textbf{Rollout time at every data budget.} Table~\ref{tab:test-time-budgets} gives the values of the curves on the right of Figure~\ref{fig:scaling-budgets}. At 1.3, 2.6, and 5.2 hours of data, \methodname{} succeeds more often than standard demonstrations at every rollout limit from one to six minutes. At 0.65 hours, standard demonstrations succeed more often at every rollout limit. The packing ratio under the six-minute limit follows the same order at each data budget. At 5.2 hours, it reaches 82.8\% for \methodname{}, compared with 76.8\% for standard demonstrations.

\begin{table}[h]
\centering
\small
\setlength{\tabcolsep}{3.5pt}
\renewcommand{\arraystretch}{1.12}
\begin{tabular}{@{}llrrrrrr@{}}
\toprule
& & \multicolumn{3}{c}{Success (\%) by limit} & \shortstack{Packing\\ratio (\%)} & \shortstack{Time used\\(min)} & \shortstack{Late\\completion (\%)} \\
\cmidrule(lr){3-5}
Data & Method & 2 min & 4 min & 6 min & & & \\
\midrule
0.65\,h & Standard & 14.0 & 15.8 & 16.5 & 33.7 & 5.20 & 0.8 \\
0.65\,h & \methodname{} & 4.7 & 5.7 & 6.0 & 27.8 & 5.74 & 0.4 \\
\addlinespace[3pt]
1.3\,h & Standard & 21.3 & 25.3 & 26.8 & 46.8 & 4.73 & 2.0 \\
1.3\,h & \methodname{} & 32.8 & 40.2 & 43.2 & 64.7 & 4.00 & 5.0 \\
\addlinespace[3pt]
2.6\,h & Standard & 36.3 & 44.0 & 46.5 & 65.8 & 3.88 & 4.5 \\
2.6\,h & \methodname{} & 49.7 & 58.0 & 60.3 & 74.9 & 3.12 & 5.6 \\
\addlinespace[3pt]
5.2\,h & Standard & 47.8 & 56.7 & 58.7 & 76.8 & 3.28 & 4.6 \\
5.2\,h & \methodname{} & 55.5 & 65.2 & 69.3 & 82.8 & 2.79 & 12.0 \\
\bottomrule
\end{tabular}
\caption{\textbf{Rollout-time statistics at matched data budgets.} Standard denotes the policies trained on standard demonstrations. Packing ratio and mean time used are measured under the six-minute limit, with failed trials counted at the limit for time used. Late completion is the percentage of trials still running at four minutes that succeed by six minutes.}
\label{tab:test-time-budgets}
\end{table}

\section{Online finetuning protocol and task-level results}
\label{app:online-task-curves}

This appendix gives the details of the online finetuning in Section~\ref{sec:online-finetuning} and shows the results on each task. Table~\ref{tab:training-settings} lists the settings, and Appendix~\ref{app:real-online} describes online finetuning on the real robot.

In simulation, each round of filtered BC (Section~\ref{sec:prelim}) admits one successful trajectory ($n_s=1$) and then performs ten gradient updates. We run fifty rounds, so each run admits fifty successful trajectories and performs 500 gradient updates.

\textbf{Training batches and replay.}
Each gradient update uses 256 training examples: 128 from offline demonstrations and 128 from admitted online trajectories.
In the first round, all 128 online examples come from the newly admitted trajectory.
In later rounds, half of the online examples come from the newly admitted trajectory and half from the buffer of earlier online successes.
Earlier trajectories are sampled uniformly with replacement, so each of the $m$ trajectories in this buffer is selected with probability $1/m$.

\textbf{Optimization.} Table~\ref{tab:training-settings} lists the learning rate, the averaging decay, and the criterion for the timesteps sampled from online trajectories.

\begin{figure}[tp]
\centering
\includegraphics[width=0.6305\linewidth,height=0.22\textheight,keepaspectratio]{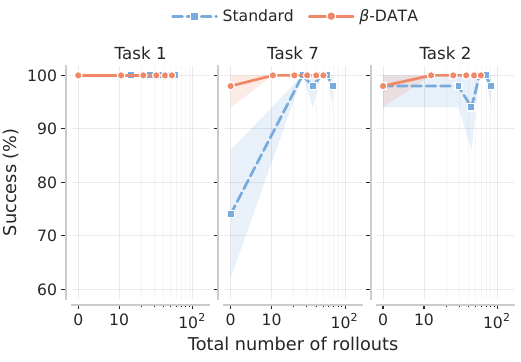}\par\smallskip
\includegraphics[width=0.8153\linewidth,height=0.22\textheight,keepaspectratio]{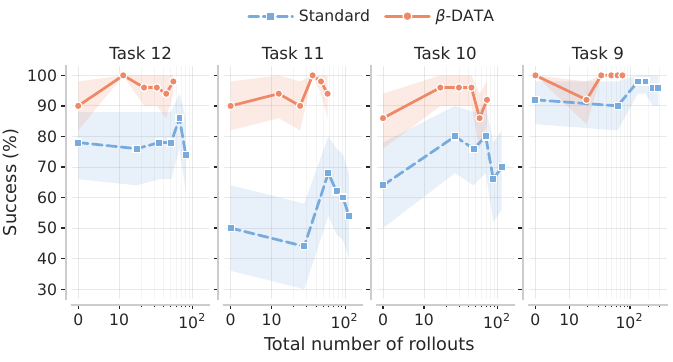}\par\smallskip
\includegraphics[width=\linewidth,height=0.22\textheight,keepaspectratio]{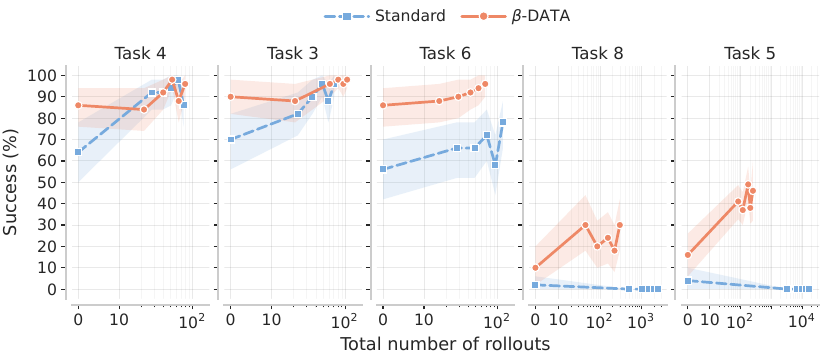}
\caption{\footnotesize \textbf{Online finetuning on individual OOD tasks.} Rows show Short, Medium, and Long tasks. Each panel shows one task. The horizontal axis of each panel ends at the larger of the final rollout counts of the two methods, and the tasks in each row are sorted by this count, from fewest to most rollouts. The horizontal axes are linear from 0 to 10 rollouts and logarithmic above 10. Shading shows pointwise 95\% bootstrap intervals for success. Standard is blue and dashed, and \methodname{} is orange and solid.}
\label{fig:online-ood-tasks}
\end{figure}

\begin{figure}[tp]
\centering
\includegraphics[width=0.6305\linewidth]{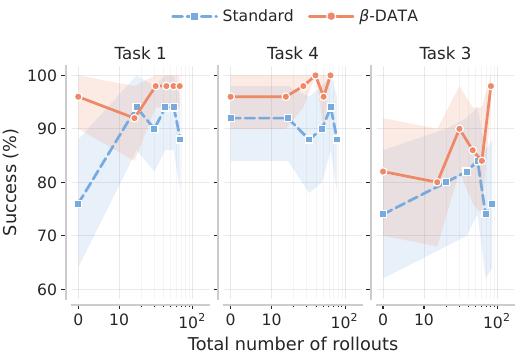}\par\smallskip
\includegraphics[width=\linewidth]{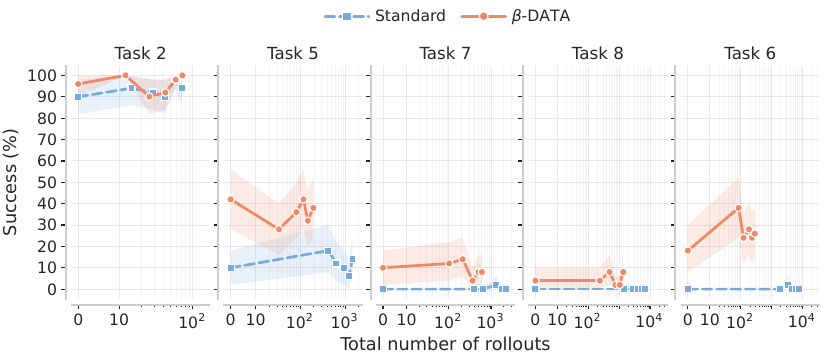}
\caption{\footnotesize \textbf{Online finetuning on individual ID tasks.} Rows show Medium and Long tasks. Each panel shows one task, using the ordering, axis scales, and pointwise 95\% bootstrap intervals of Figure~\ref{fig:online-ood-tasks}. Standard is blue and dashed, and \methodname{} is orange and solid.}
\label{fig:online-id-tasks}
\end{figure}

\textbf{Per-task results.} Figure~\ref{fig:online-ood-tasks} shows the success rate of online finetuning on each OOD task as a function of the number of rollouts, for policies midtrained on standard demonstrations and on \methodname{}. Figure~\ref{fig:online-id-tasks} shows the same curves for the ID tasks.

\section{Connected behavioral coverage (CBC) measurement}
\label{app:cbc-measurement}

This appendix describes how we compute the edge counts reported in Section~\ref{sec:method} and plotted in Figure~\ref{fig:cbc-budget}. Table~\ref{tab:cbc-inventory} lists the annotated demonstrations and the resulting counts for each domain.

\textbf{Counting directed edges.} Each edge comes from one annotated trajectory segment and goes from its starting abstract state to its ending abstract state. For both datasets of a domain, we use the representation $\phi(s,o_i)$ from Section~\ref{sec:method}, which maps the tracked object and its location to an abstract state. An abstract state does not include the pose of the object, the rest of the scene, or the task goal. As a result, two segments that start and end in the same abstract states count as one edge, even when their low-level action sequences differ.

\textbf{Annotated demonstrations.} We annotate the standard demonstrations and the \methodname{} demonstrations for simulated packing, real-world packing and LEGO disassembly, and count the distinct directed edges (Eq.~\ref{eq:cbc}) in each dataset. \methodname{} contains more distinct directed edges than the standard demonstrations. For each dataset, Table~\ref{tab:cbc-inventory} lists the number of hours, the number of demonstrations, and the number of distinct starting and ending states when all of its demonstrations are used. A starting state is the abstract state at the beginning of an annotated segment, and an ending state is the abstract state at its end. The same abstract state can appear both as a starting state and as an ending state.

\textbf{Matched data budgets.} To select the demonstrations under the matched budget, we shuffle the demonstrations of a dataset and keep adding whole demonstrations in that order for as long as they fit within the budget. We repeat this selection for 1,000 random orders and report the median edge count.

\begin{table}[h]
\centering
\small
\begin{tabular}{@{}llrrrrr@{}}
\toprule
Domain & Data & Hours & \shortstack{Demon-\\strations} & \shortstack{Edges\\(matched\\budget)} & \shortstack{Starting\\states} & \shortstack{Ending\\states} \\
\midrule
Simulated packing & Standard & 6.1 & 288 & 54 & 27 & 35 \\
 & \methodname{} & 5.9 & 200 & 85 & 36 & 37 \\
\addlinespace
Real-world packing & Standard & 4.4 & 312 & 72 & 36 & 68 \\
 & \methodname{} & 4.9 & 156 & 125 & 65 & 76 \\
\addlinespace
LEGO disassembly & Standard & 4.0 & 103 & 15 & 9 & 13 \\
 & \methodname{} & 4.0 & 54 & 19 & 13 & 13 \\
\bottomrule
\end{tabular}
\caption{\footnotesize{\textbf{Annotated demonstrations used to measure CBC.} Standard denotes standard demonstrations. Hours, demonstrations, starting states and ending states are computed using all demonstrations of a dataset. Edges (matched budget) is the median number of distinct directed edges (Eq.~\ref{eq:cbc}) over 1,000 random selections at the matched data budget of the domain.}}
\label{tab:cbc-inventory}
\end{table}

\textbf{CBC with pretrained visual features.} The bottom-right panel of Figure~\ref{fig:cbc-latent} measures CBC in simulated packing with abstract states computed from pretrained visual features. For each annotated segment, we take the pretrained $\pi_{0.5}$ vision features of the top-camera image at both endpoints, averaged over image tokens. When the adversary changes the scene during a segment, the start is the first observation after the last change, so the transition begins from the scene the operator faces. We cluster the endpoint features of both datasets with $k$-means, and a segment contributes a directed edge when it starts and ends in different clusters. We compare the datasets at the same recorded hours, drawing standard demonstrations to match the hours of \methodname{}, and report the interquartile mean over clustering seeds and draws with bootstrap intervals. \methodname{} also contains more distinct transitions when both datasets contribute the same number of segments, though the difference is smaller.

\section{Simulated packing data collection and policy rollouts}
\label{app:sim-packing-filmstrips}
\begin{figure}[h]
\centering
\includegraphics[width=\linewidth]{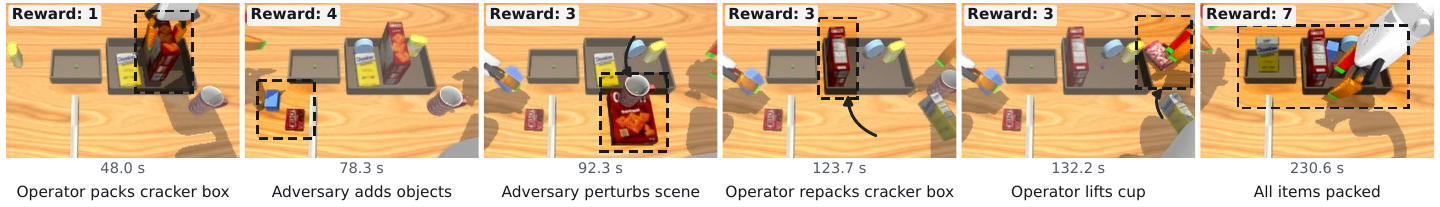}
\caption{\footnotesize \textbf{Data collection in simulated packing.} Simulated perturbations (such as adding objects to the task or moving packed objects back onto the table) are applied and the human operator continues packing. The operator packs the cracker box again and lifts the cup, and then completes the goal. Dashed boxes show the relevant objects. Curved arrows show the direction of movement.}
\label{fig:sim-protocol-strip}
\end{figure}

\begin{figure}[h]
\centering
\vspace{-0.2cm}
\includegraphics[width=0.99\linewidth]{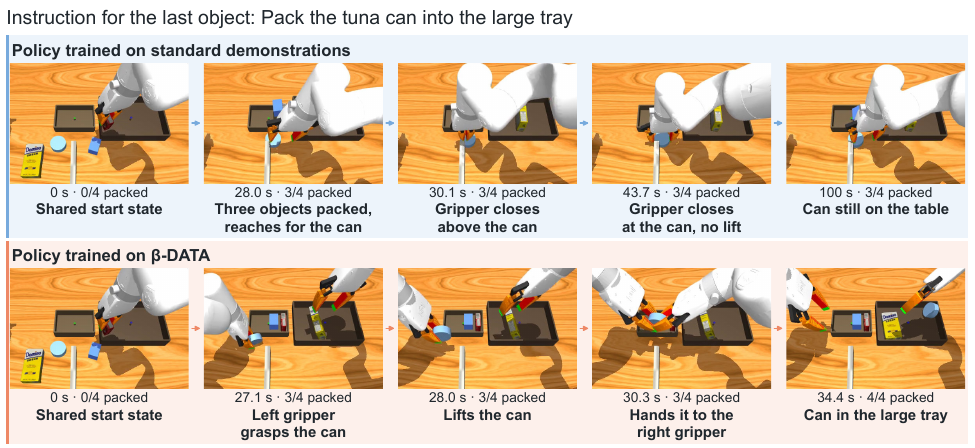}
\vspace{-0.3cm}
\caption{\footnotesize{\textbf{Simulated packing rollouts of two policies from the same initial state.} The two rows are selected paired rollouts from a separate comparison, in which the policy must pack four objects. Top row: the policy trained on standard demonstrations packs three objects and is then instructed to pack the tuna can. It closes its gripper 23 times near the tuna can and never lifts it, and the tuna can is still on the table when the rollout ends. Bottom row: the policy trained on \methodname{} grasps the tuna can with its left gripper, lifts it, hands it to its right gripper, and packs it into the large tray. It completes the four-object task in 34.4\,s. The text below each frame gives the execution time and the number of packed objects.}}
\label{fig:packing-pickup}
\vspace{-0.0cm}
\end{figure}

\begin{figure}[t]
\centering
\vspace{-0.2cm}
\includegraphics[width=0.99\linewidth]{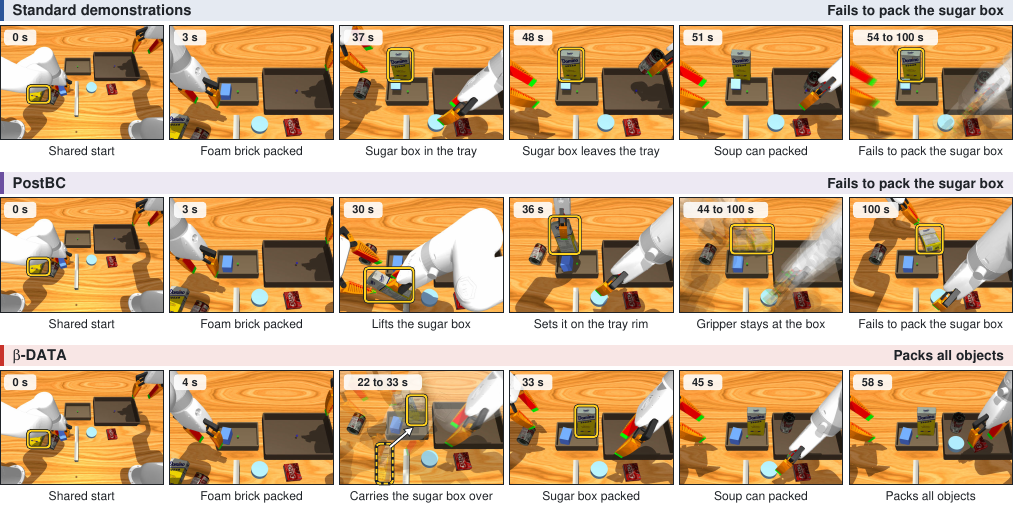}
\vspace{-0.25cm}
\caption{\footnotesize{\textbf{From the same initial state, only the policy trained on \methodname{} packs the sugar box in these rollouts.} The three policies start from the state of Figure~\ref{fig:sim-task11-recovery} and are instructed to pack the sugar box into the medium tray. Yellow outlines mark the sugar box. Frames with a time range overlay several frames from that range. The policy trained on standard demonstrations (top) places the sugar box, but the box leaves the tray, and the policy does not return to it. PostBC (middle) lifts the sugar box and sets it on the tray rim, and its gripper stays at the box until the rollout ends. The policy trained on \methodname{} (bottom) carries the sugar box into the tray and then packs the remaining objects.}}
\label{fig:sim-task11-postbc}
\vspace{-0.3cm}
\end{figure}

This appendix gives simulated packing examples for the data collection protocol in Section~\ref{sec:method} and for the zero-shot results in Section~\ref{sec:zero-shot-tasks}.
Figure~\ref{fig:sim-protocol-strip} shows how the operator continues the task when perturbations change the scene during data collection.
Figure~\ref{fig:packing-pickup} compares a policy trained on standard demonstrations with a policy trained on \methodname{}, starting from the same initial state.
Figure~\ref{fig:sim-task11-postbc} compares these two policies and PostBC from the state of Figure~\ref{fig:sim-task11-recovery}.

\FloatBarrier
\section{Real-world packing data collection and policy rollouts}
\label{app:real-xarm-rollouts}
\label{app:real-online}

\begin{figure}[htbp]
\centering
\includegraphics[width=\linewidth]{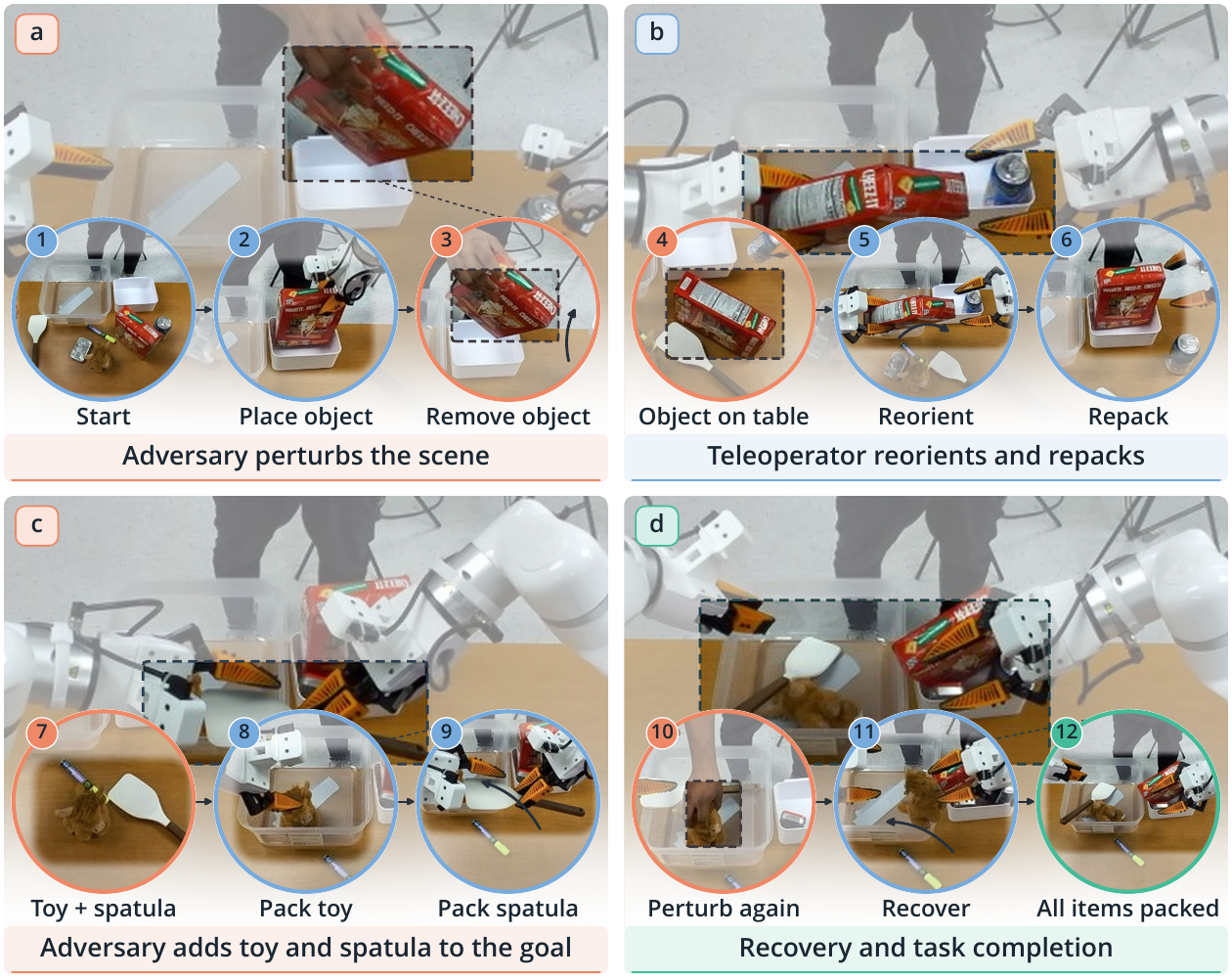}
\caption{\footnotesize \textbf{Perturbation and recovery during real-world data collection.} In one teleoperated episode, the adversary removes packed objects and adds the toy and spatula to the goal. The operator recovers and completes the updated packing task. Numbered circles show selected moments in chronological order.}
\label{fig:real-collection-recovery}
\end{figure}

\textbf{Adversarial data collection.}
Figure~\ref{fig:real-collection-recovery} shows a sample teleoperated episode collected with the adversarial data collection protocol from Section~\ref{sec:method}. 

\begin{figure}[htbp]
\centering
\includegraphics[width=\linewidth]{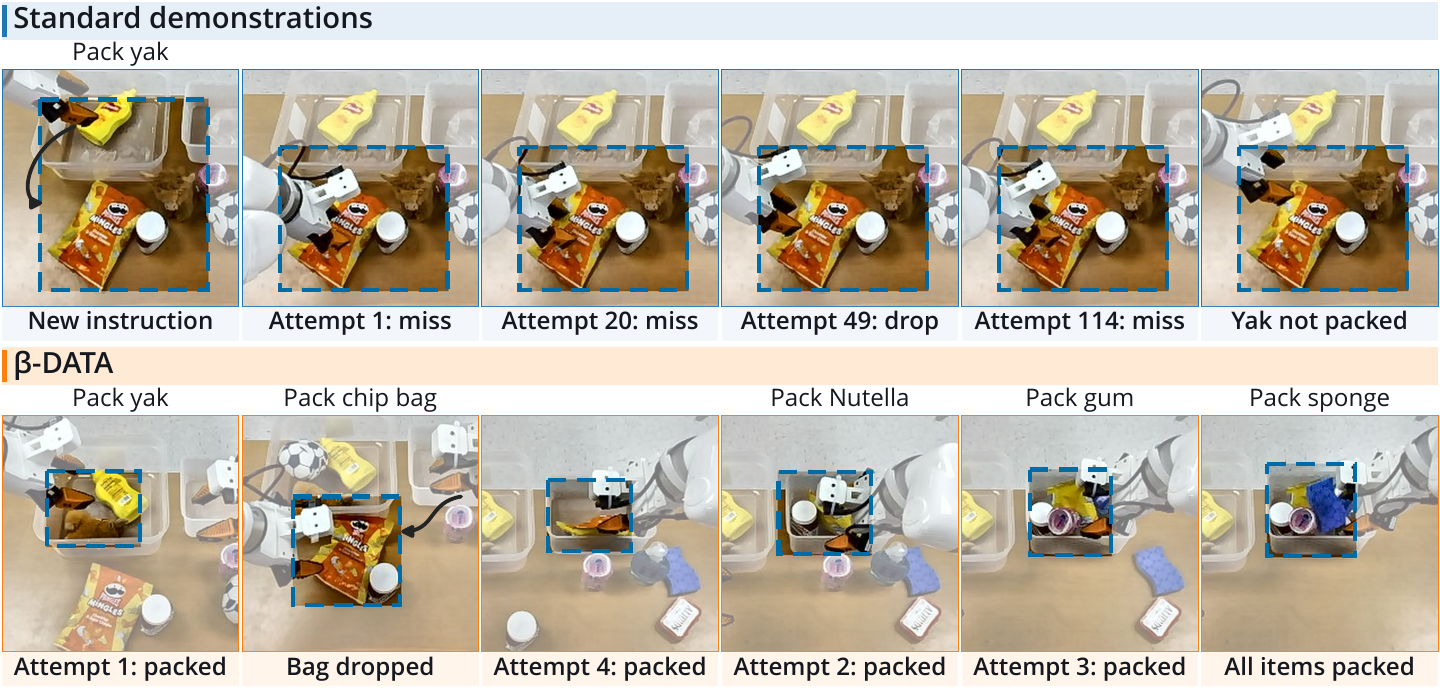}
\caption{\footnotesize \textbf{Zero-shot rollouts on real-world xArm packing.} The policy trained on standard demonstrations (top) makes repeated grasp attempts on the chip bag while it is instructed to pack the yak, and the yak is not packed at the end. The policy trained on \methodname{} (bottom) packs the yak on its first attempt, recovers after the chip bag drops, and completes the task. Dashed boxes identify relevant objects, and arrows indicate schematic movement directions.}
\label{fig:real-xarm-head-to-head}
\end{figure}

\begin{figure}[htbp]
\centering
\includegraphics[width=\linewidth]{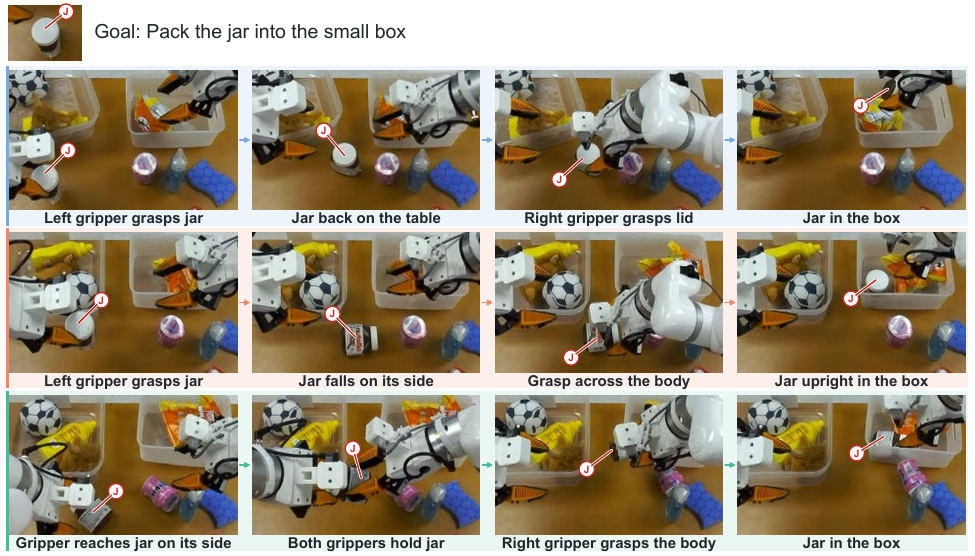}
\caption{\footnotesize{\textbf{Under one instruction, the policy trained on \methodname{} packs a jar through three different sequences of grasps in real-world packing.} The pin marks the jar in every frame. In the top and middle rollouts, the jar starts upright on the table and the left gripper grasps it. In the top rollout, the jar ends up back on the table, and the right gripper grasps its lid and packs it. In the middle rollout, the jar falls on its side, and the right gripper grasps it across its body and places it upright in the box. In the bottom rollout, the jar starts on its side. The policy holds the jar with both grippers and, after further attempts, grasps it across its body with the right gripper and packs it.}}
\label{fig:real-xarm-jar-routes}
\end{figure}

\textbf{Zero-shot rollouts.}
Figure~\ref{fig:real-xarm-head-to-head} compares zero-shot rollouts of the policies trained on standard demonstrations and on \methodname{}. The policy trained on \methodname{} recovers after the chip bag drops and completes the~task. As Figure~\ref{fig:lego-removal-routes} does for LEGO disassembly, Figure~\ref{fig:real-xarm-jar-routes} presents the connected behavioral coverage induced by training on \methodname{} in real-world packing.

\begin{figure}[htbp]
\centering
\includegraphics[width=\linewidth]{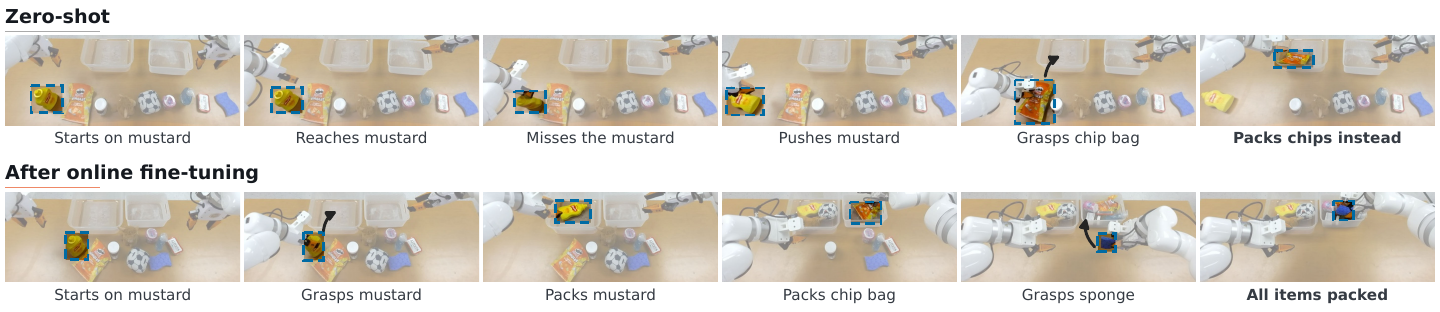}
\caption{\footnotesize \textbf{Real-world xArm packing before and after online finetuning.} Both rows show the policy trained on \methodname{}. The zero-shot policy (top) fails to grasp the mustard and packs the chip bag while it is instructed to pack the mustard. After online finetuning, the policy (bottom) packs the mustard and continues until all items are packed.}
\label{fig:real-xarm-before-after}
\end{figure}

\textbf{Before and after online finetuning.}
Figure~\ref{fig:real-xarm-before-after} compares sample rollouts of the policy trained on \methodname{} at the midtraining checkpoint and after online finetuning. Table~\ref{tab:real-online-results} reports the success rates of these two checkpoints.

\Needspace{7\baselineskip}\textbf{Online finetuning settings.}
In real-world packing, starting from the midtraining checkpoint of each policy, each run collects twenty admitted successful trajectories and performs ten gradient updates after each admitted success, for 200 updates in total.
Each batch has 256 examples: 128 come from offline demonstrations and 128 from admitted online trajectories.
The learning rate increases to $3\times10^{-5}$ over ten warmup updates and remains constant thereafter.

\FloatBarrier
\section{LEGO data collection and policy rollouts}
\label{app:lego-filmstrips}
\begin{figure}[h]
\centering
\includegraphics[width=0.925\linewidth]{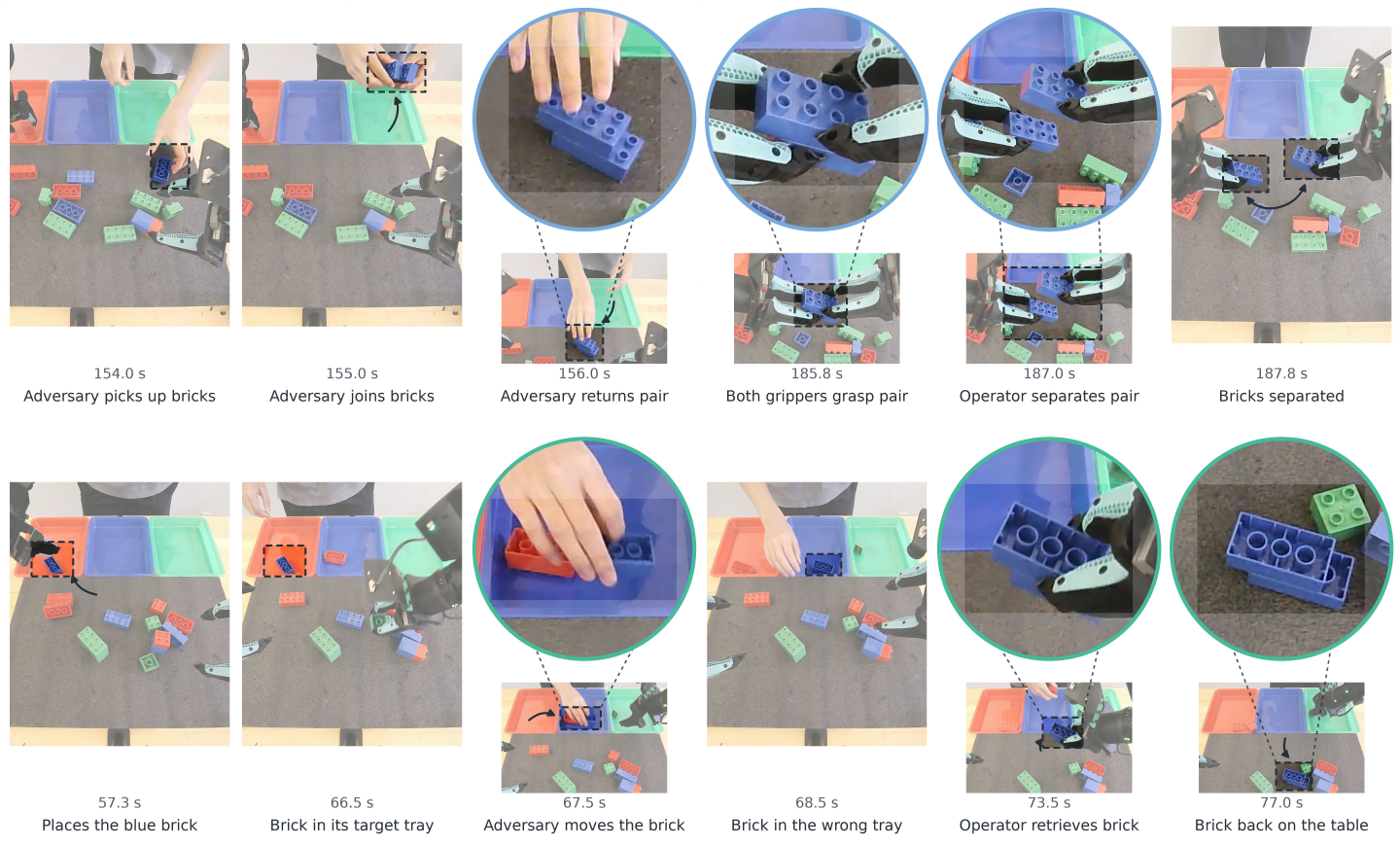}
\caption{\footnotesize \textbf{Data collection in LEGO disassembly and sorting.} The two rows show two separate sequences. Top row: the adversary joins two bricks, and the operator separates them with both arms. Bottom row: the adversary moves a placed brick into another tray, and the operator takes it back. Dashed boxes show the relevant bricks. Arrows show the direction of~movement.}
\label{fig:lego-collection}
\end{figure}

\begin{figure}[h]
\centering
\includegraphics[width=0.925\linewidth]{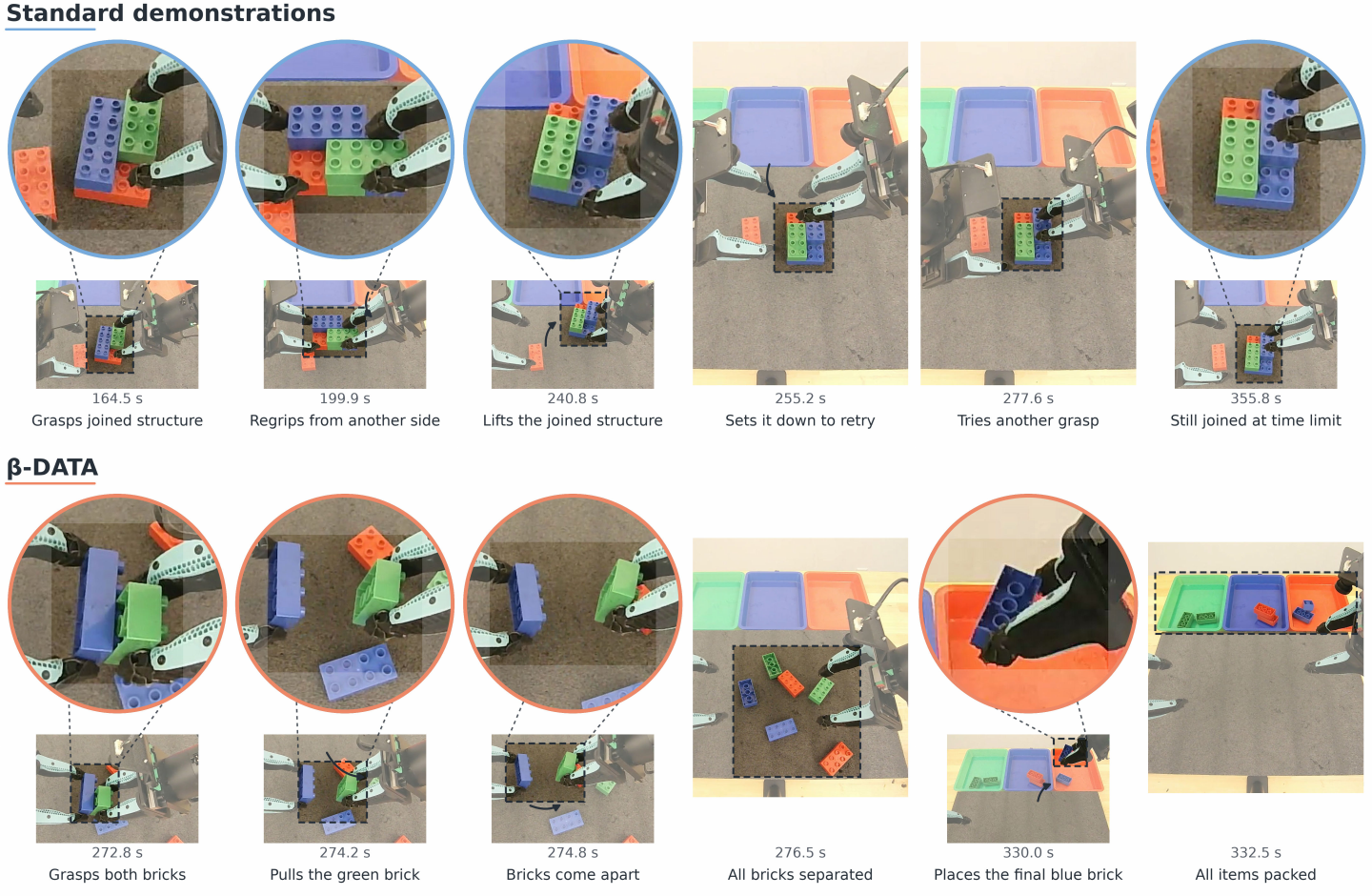}
\caption{\footnotesize \textbf{LEGO disassembly and sorting rollouts.} Top row: the policy trained on standard demonstrations regrasps the structure many times but does not separate its bricks. Bottom row: the policy trained on \methodname{} separates the bricks and places the final brick to complete sorting. Dashed boxes show the relevant bricks.}
\label{fig:lego-policy-rollouts}
\end{figure}

Figure~\ref{fig:lego-collection} shows how the operator responds when the adversary rejoins or relocates bricks during collection.
Figure~\ref{fig:lego-policy-rollouts} illustrates the policies' different grasping and disassembly behavior in selected evaluation episodes.

\section{Frontier models as robot policies}
\label{app:frontier}

Our tasks require low-level behaviors that are difficult to specify in language (Section~\ref{sec:intro}). If language-level reasoning were sufficient, a frontier model that plans well should also be able to carry out these tasks. We test this directly: we run GPT-6.2 Astra as the robot policy on real-world LEGO disassembly and sorting, with no task-specific training, and report the result in Section~\ref{sec:zero-shot-tasks}. This appendix gives the details.

\textbf{Setup.} We use GPT-6.2 Astra as the policy for the YAM arms. At each step, the model receives the overhead camera image and both wrist camera images, the measured end-effector pose and gripper opening of each arm, its recent actions, and the current subtask instruction. As in-context demonstrations, the prompt also includes overhead camera keyframes from teleoperated demonstrations of the same subtask, which show the model how a human operator holds the structure and removes each brick. The model outputs either a single-arm action, specified by a change in end-effector position and orientation relative to the measured pose and a gripper command, or a subtask completion signal. The other arm maintains its pose and grip, so one arm can hold the structure while the other pulls a brick. We ask for a change rather than an absolute target because the model estimates relative motion from images more reliably than an exact position and orientation. A controller adds the change to the measured pose, solves inverse kinematics, and moves the end effector within speed limits. It rejects moves that leave the workspace, have no inverse kinematics solution, or would cause the arms to collide, and reports the reason to the model at the next step. The model has no tools or code execution. A high-level planner provides the current subtask instruction.

\begin{figure}[h]
\centering
\includegraphics[width=\linewidth]{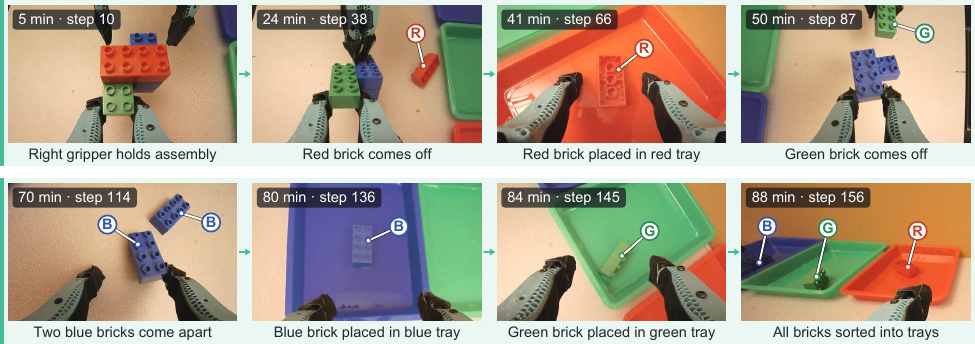}
\caption{\footnotesize \textbf{GPT-6.2 Astra takes apart and sorts a single assembly in about 88 minutes.} Each frame shows the elapsed time of model control and the model step. The model separates and places one brick at a time.}
\label{fig:astra-success}
\end{figure}

\begin{figure}[h]
\centering
\includegraphics[width=\linewidth]{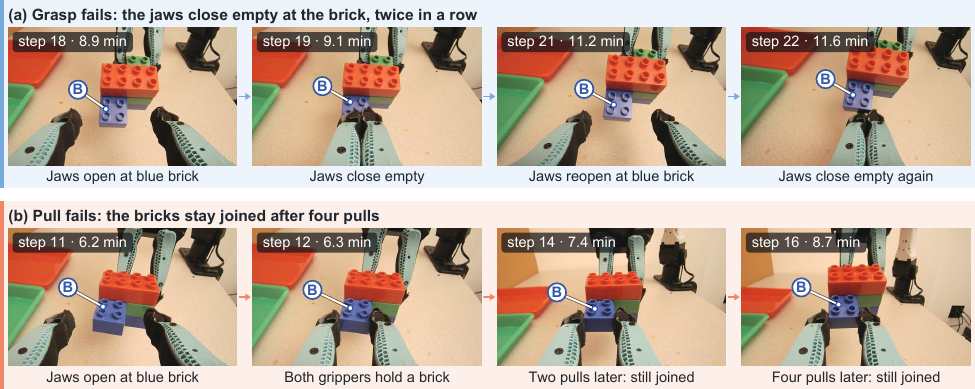}
\caption{\footnotesize \textbf{GPT-6.2 Astra follows a correct plan but fails at the low-level execution.} Left wrist camera view. Each frame shows the model step and elapsed time. (a) The left gripper fails to grasp the blue brick on two attempts despite reopening and repositioning. (b) Both grippers grasp a brick, but fail to pull the bricks apart even after multiple attempts.}
\label{fig:astra-failure}
\end{figure}

\textbf{Results.} Across nine runs, the model successfully completes the task once, taking 88 minutes, of which 81 minutes are spent waiting for model outputs, averaging 31 seconds per step (Figure~\ref{fig:astra-success}). The model follows a consistent strategy: it holds the structure with one arm, pulls a brick with the other, checks the wrist camera image, and retries after failed attempts. It also adapts to the scene. When the controller rejects a motion that would bring the grippers too close, the model moves one arm out of the way and retries. After grasping the red and green bricks with separate grippers, it identifies the blue brick between them as the next to remove. Once a brick is free, it places the brick in its tray. However, two failures recur during execution (Figure~\ref{fig:astra-failure}). First, the gripper closes beside or short of the brick, and the grasp fails again and again. Second, when holding a brick in each gripper, the model rocks, twists, or pulls at an angle rather than pulling the bricks apart. These motions tilt or shift the joined bricks or cause a grasp to slip, and the bricks stay joined after multiple attempts (Figure~\ref{fig:astra-failure}(b)).

\textbf{Discussion.} These runs show the gap that motivates \methodname{}: the model specifies what should happen next, but fails at simple low-level behaviors that realize it, such as aligning the gripper with a brick before closing it or pulling bricks apart. A correct plan can be realized through many different grasps, orientations, and contacts (Figure~\ref{fig:lego-removal-routes}), so a policy needs CBC at the low level. The runs thus support the validity of our tasks as tests of low-level behavior. We view the result as complementary to our study rather than in tension with it. \methodname{} concerns the data from which a policy learns low-level behavior, and a frontier model that acts through low-level actions would face the same question of which behaviors its training data covers.

\end{document}